\documentclass[runningheads]{llncs}
\usepackage{graphicx}
\usepackage{amsmath,amssymb} % define this before the line numbering.
\usepackage{color}
\usepackage{epsfig}
\usepackage{epstopdf}
\usepackage{booktabs}
\usepackage{wrapfig}
\usepackage{sidecap}
\usepackage{colortbl}
\definecolor{Gray}{gray}{0.8}
\usepackage[ruled,vlined]{algorithm2e}
\newcommand{\etal}{\textit{et al}.}

\usepackage{nonfloat}
\begin{document}
\title{ICNet for Real-Time Semantic Segmentation\\ on High-Resolution Images}
% Replace with your title

\titlerunning{ICNet for Real-Time Semantic Segmentation}
% Replace with a meaningful short version of your title
%
%\author{First Author\inst{1}\orcidID{0000-1111-2222-3333} \and
%Second Author\inst{2,3}\orcidID{1111-2222-3333-4444} \and
%Third Author\inst{3}\orcidID{2222--3333-4444-5555}}
\author{Hengshuang Zhao$^{1}$, Xiaojuan Qi$^{1}$, Xiaoyong Shen$^2$, Jianping Shi$^3$, Jiaya Jia$^{1,2}$}
%
%Please write out author names in full in the paper, i.e. full given and family names. 
%If any authors have names that can be parsed into FirstName LastName in multiple ways, please include the correct parsing, in a comment to the volume editors:
%\index{Lastnames, Firstnames}
%(Do not uncomment it, because you may introduce extra index items if you do that, we will use scripts for introducing index entries...)
\authorrunning{H. Zhao, X. Qi, X. Shen, J. Shi, J. Jia}
% Replace with shorter version of the author list. If there are more authors than fits a line, please use A. Author et al.
%

\institute{$^1$The Chinese University of Hong Kong, $^2$ Tencent Youtu Lab, $^3$SenseTime Research
%\email{\scriptsize{\{hszhao,xjqi,leojia\}@cse.cuhk.edu.hk, dylanshen@tencent.com, shijianping@sensetime.com}}
\email{\{hszhao,xjqi,leojia\}@cse.cuhk.edu.hk, \\dylanshen@tencent.com, shijianping@sensetime.com}
}

\maketitle              % typeset the header of the contribution
\setcounter{footnote}{-1}
\begin{abstract}
We focus on the challenging task of real-time semantic segmentation in this paper. It finds many practical applications and yet is with fundamental difficulty of reducing a large portion of computation for pixel-wise label inference. We propose an image cascade network (ICNet) that incorporates multi-resolution branches under proper label guidance to address this challenge. We provide in-depth analysis of our framework and introduce the cascade feature fusion unit to quickly achieve high-quality segmentation. Our system yields real-time inference on a single GPU card with decent quality results evaluated on challenging datasets like Cityscapes, CamVid and COCO-Stuff.
\end{abstract}
\section{Introduction}
Semantic image segmentation is a fundamental task in computer vision. It predicts dense labels for all pixels in the image, and is regarded as a very important task that can help deep understanding of scene, objects, and human. Development of recent deep {\it convolutional neural networks} (CNNs) makes remarkable progress on semantic segmentation~\cite{long2015fully,chen2015semantic,badrinarayanan2015segnet,noh2015learning,zhao2017pspnet,wu2016wider}. The effectiveness of these networks largely depends on the sophisticated model design regarding depth and width, which has to involve many operations and parameters.

CNN-based semantic segmentation mainly exploits \textit{fully convolutional networks} (FCNs). It is common wisdom now that increase of result accuracy almost means more operations, especially for pixel-level prediction tasks like semantic segmentation. To illustrate it, we show in Fig.~\ref{fig:accuracytime-timeimageresolution}(a) the accuracy and inference time of different frameworks on Cityscapes~\cite{cordts2016cityscapes} dataset.

\subsubsection{Status of Fast Semantic Segmentation}
Contrary to the extraordinary development of high-quality semantic segmentation, research along the line to make semantic segmentation run {\it fast} while not sacrificing too much quality is left behind. We note actually this line of work is similarly important since it can inspire or enable many practical tasks in, for example, automatic driving, robotic interaction, online video processing, and even mobile computing where running time becomes a critical factor to evaluate system performance.

\begin{figure}[t]
	\begin{minipage}[t]{0.43\linewidth}
		\centering
		\includegraphics[width=0.94\linewidth]{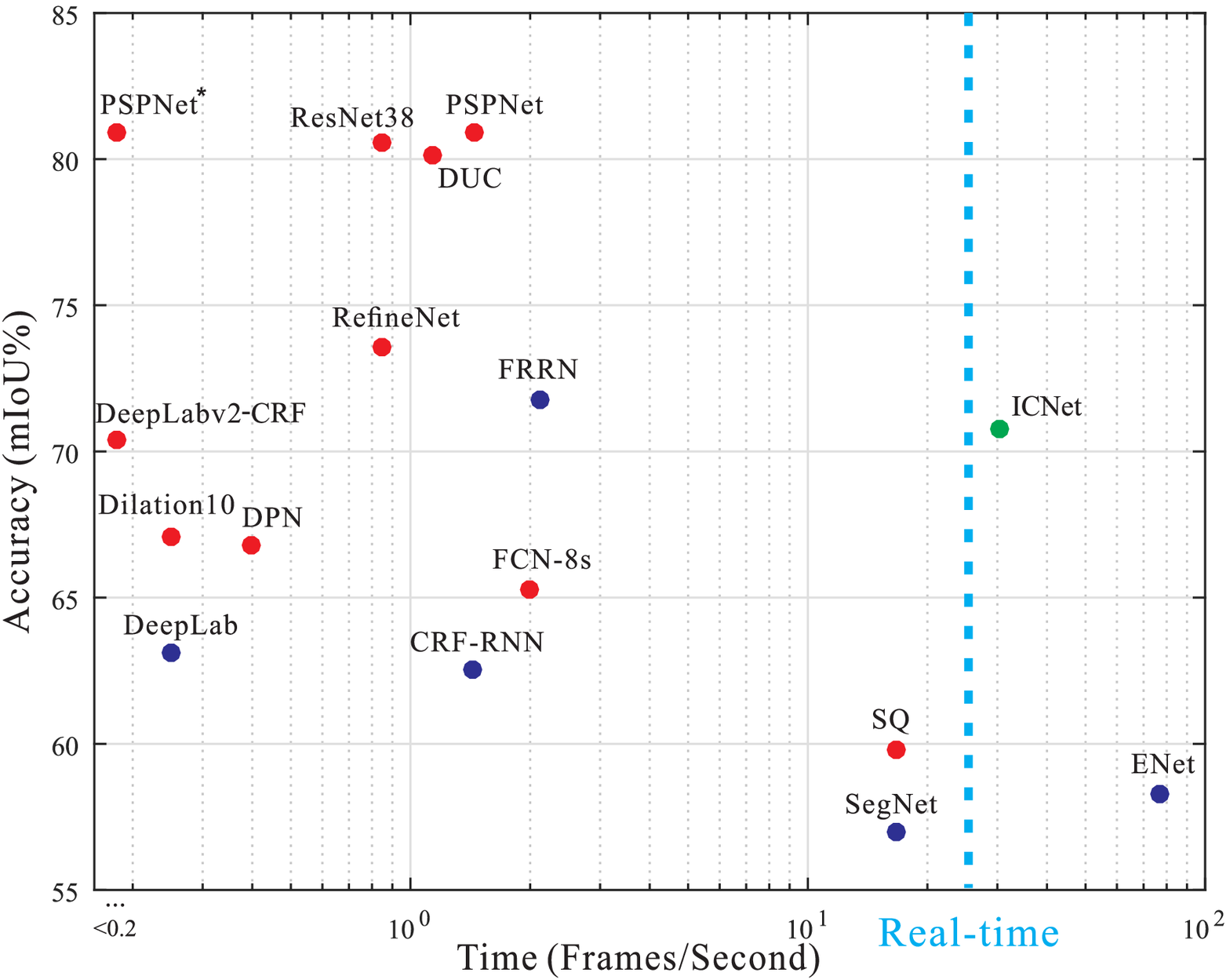}
		{\scriptsize \\(a) Inference speed and mIoU}
	\end{minipage}
	\begin{minipage}[t]{0.55\linewidth}
		\centering
		\includegraphics[width=1.0\linewidth]{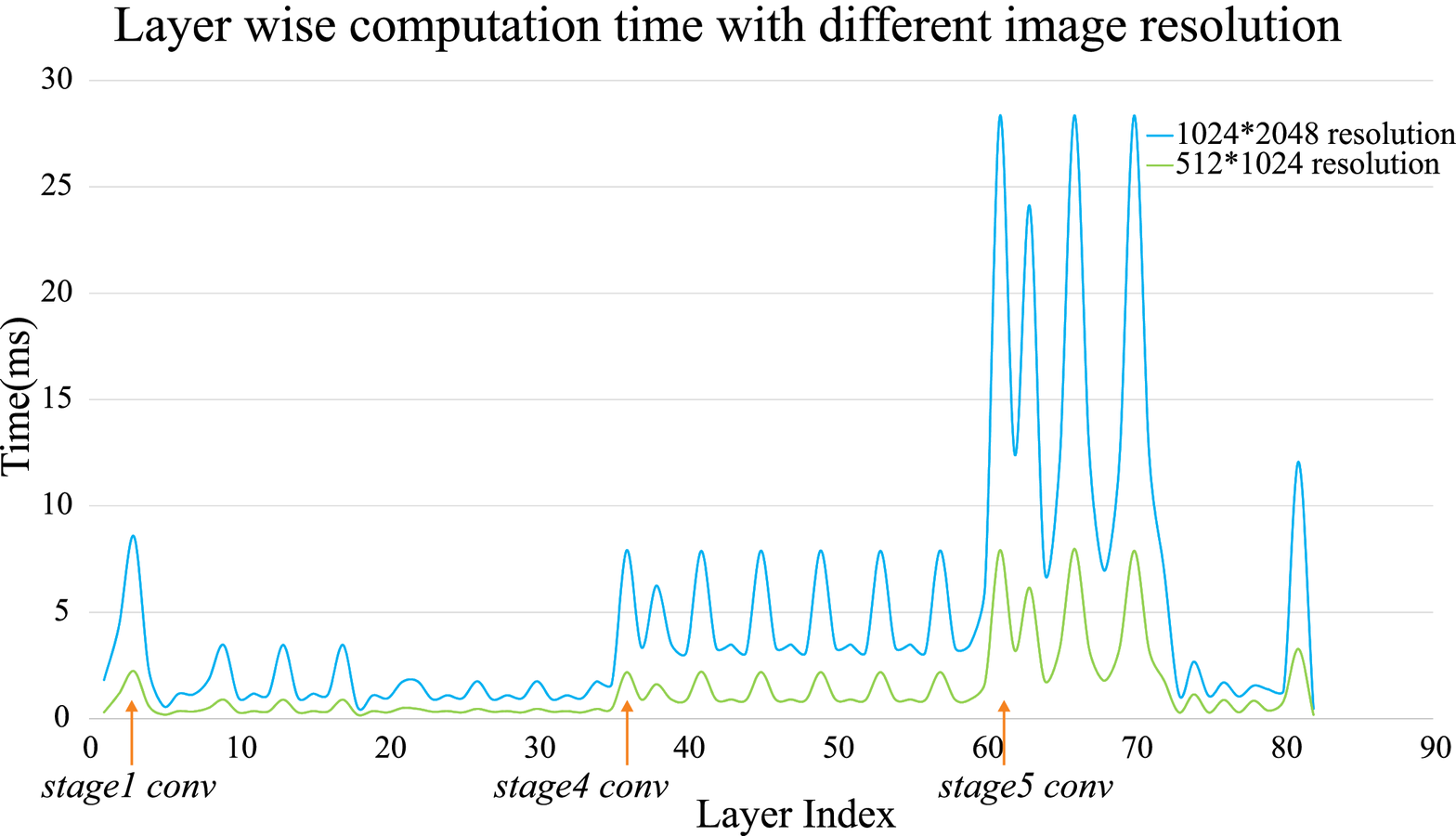}
		{\scriptsize (b) Time in each layer of PSPNet50}
	\end{minipage}
	\caption{\textbf{(a)\protect\footnotemark:} Inference speed and mIoU performance on Cityscapes~\cite{cordts2016cityscapes}
		test set. Methods involved are PSPNet~\cite{zhao2017pspnet}, ResNet38~\cite{wu2016wider},
		DUC~\cite{wang2017duc}, RefineNet~\cite{lin2017refine}, %LRR~\cite{ghiasi2016laplacian},
		FRRN~\cite{pohlen2017FRRN}, DeepLabv2-CRF\cite{chen2016deeplab}, Dilation10~\cite{yu2016multi},
		DPN~\cite{liu2015semantic}, FCN-8s~\cite{long2015fully}, DeepLab~\cite{chen2015semantic},
		CRF-RNN~\cite{zheng2015conditional}, SQ~\cite{treml2016speeding},
		ENet~\cite{paszke2016enet}, SegNet~\cite{badrinarayanan2015segnet}, and our ICNet.
		\textbf{(b):} Time spent on PSPNet50 with dilation 8 for two input images. Roughly running time is proportional to the pixel number and kernel number.}
	\label{fig:accuracytime-timeimageresolution}
\end{figure}

Our experiments show that high-accuracy methods of ResNet38~\cite{wu2016wider} and PSPNet~\cite{zhao2017pspnet} take around 1 second to predict a $1024\times 2048$ high-resolution image on one Nvidia TitanX GPU card during testing. These methods fall into the area illustrated in Fig. \ref{fig:accuracytime-timeimageresolution}(a) with high accuracy and low speed. Recent fast semantic segmentation methods of ENet~\cite{paszke2016enet} and SQ~\cite{treml2016speeding}, contrarily, take quite different positions in the plot. The speed is much accelerated; but accuracy drops, where the final mIoUs are lower than 60\%. These methods are located in the lower right phase in the figure.

\footnotetext{Blue ones are tested with downsampled images. Inference speed is reported with single network forward while accuracy of several mIoU aimed approaches (like PSPNet$^\star$) may contain testing tricks like multi-scale and flipping, resulting much more time. See supplementary material for detailed information.}

\subsubsection{Our Focus and Contributions}
In this paper, we focus on building a practically fast semantic segmentation system with decent prediction accuracy. Our method is the first in its kind to locate in the top-right area shown in Fig.~\ref{fig:accuracytime-timeimageresolution}(a) and is one of the only two available real-time approaches. It achieves decent trade-off between efficiency and accuracy.

Different from previous architectures, we make comprehensive consideration on the two factors of speed and accuracy that are seemingly contracting. We first make in-depth analysis of time budget in semantic segmentation frameworks and conduct extensive experiments to demonstrate insufficiency of intuitive speedup strategies. This motivates development of \textit{image cascade network} (ICNet), a high efficiency segmentation system with decent quality. It exploits efficiency of processing low-resolution images and high inference quality of high-resolution ones. The idea is to let low-resolution images go through the full semantic perception network first for a coarse prediction map. Then cascade feature fusion unit and cascade label guidance strategy are proposed to integrate medium and high resolution features, which refine the coarse semantic map gradually. We make all our code and models publicly available\footnote{\href{https://github.com/hszhao/ICNet}{https://github.com/hszhao/ICNet}}. Our main contributions and performance statistics are the following.

\begin{itemize}
	\item We develop a novel and unique image cascade network for real-time semantic segmentation, it utilizes semantic information in low resolution along with details from high-resolution images efficiently. 
	\item The developed cascade feature fusion unit together with cascade label guidance can recover and refine segmentation prediction progressively with a low computation cost.
	\item Our ICNet achieves 5$\times$ speedup of inference time, and reduces memory consumption by 5$\times$ times. It can run at high resolution $1024\times 2048$ in speed of 30 fps while accomplishing high-quality results. It yields real-time inference on various datasets including Cityscapes~\cite{cordts2016cityscapes}, CamVid~\cite{BrostowFC09} and COCO-Stuff~\cite{caesar2016coco}.
\end{itemize}

\section{Related Work}
Traditional semantic segmentation methods~\cite{liu2011nonparametric} adopt handcrafted feature to learn the representation. Recently, CNN based methods largely improve the performance.

\subsubsection{High Quality Semantic Segmentation}
FCN~\cite{long2015fully} is the pioneer work to replace the last fully-connected layers in classification with convolution layers. DeepLab~\cite{chen2015semantic,chen2016deeplab} and \cite{yu2016multi} used dilated convolution to enlarge the receptive field for dense labeling. Encoder-decoder structures~\cite{badrinarayanan2015segnet,noh2015learning} can combine the high-level semantic information from later layers with the spatial information from earlier ones. Multi-scale feature ensembles are also used in~\cite{chen2015attention,hariharan2015hypercolumns,xia2016zoom}. In \cite{chen2015semantic,liu2015semantic,zheng2015conditional}, conditional random fields (CRF) or Markov random fields (MRF) were used to model spatial relationship. Zhao~\etal~\cite{zhao2017pspnet} used pyramid pooling to aggregate global and local context information. Wu~\etal~\cite{wu2016wider} adopted a wider network to boost performance. In \cite{lin2017refine}, a multi-path refinement network combined multi-scale image features. These methods are effective, but preclude real-time inference.
%but cannot be directly applied to real-time inference.

\subsubsection{High Efficiency Semantic Segmentation}
In object detection, speed became one important factor in system design \cite{girshick2015fastrcnn,ren2015faster}. Recent Yolo~\cite{Redmon2016yolo,Redmon2017yolo2} and SSD~\cite{liu2016ssd} are representative solutions. In contrast, high speed inference in semantic segmentation is under-explored.
%SegNet~\cite{badrinarayanan2015segnet} adopts transferred pool indices to reduce layer parameters.
ENet~\cite{paszke2016enet} and ~\cite{romera2017efficient} are lightweight networks. These methods greatly raise efficiency with notably sacrificed accuracy.

\subsubsection{Video Semantic Segmentation}
Videos contain redundant information in frames, which can be utilized to reduce computation. Recent Clockwork~\cite{Shelhamer2016clockfcn} reuses feature maps given stable video input. Deep feature flow~\cite{zhu2017dff} is based on a small-scale optical flow network to propagate features from key frames to others. FSO~\cite{kundu2016feature} performs structured prediction with dense CRF applied on optimized features to get temporal consistent predictions. NetWarp~\cite{gadde2017semantic} utilizes optical flow of adjacent frames to warp internal features across time space in video sequences. We note when a good-accuracy fast image semantic-segmentation framework comes into existence, video segmentation will also be benefited.

%\subsubsection{Cascade Structures}
%Several modern segmentation frameworks incorporate cascade structures. FCN~\cite{long2015fully} and DeepLab-MSC~\cite{chen2015semantic} sum the multi-resolution prediction maps to generate the final score map. UNet~\cite{ronneberger2015unet} adopts skip connections during deconvolution to exploit middle-level features. LRR~\cite{ghiasi2016laplacian} applies a pyramid reconstruction architecture, where the feature maps are reconstructed from bottom up. RefineNet~\cite{lin2017refine} fuses multi-path feature maps from different layers using long-range residual connections.

%We note these methods focus on fusing features from different layers with input going through the whole segmentation processing branch. They all face the same problem of expensive computation given high-resolution input. Our method is with a hierarchical structure for fast semantic segmentation. It is by nature different from the multi-scale testing \cite{chen2016deeplab,lin2017refine} that boosts accuracy with each branch processed independently. Our image cascade network, on the contrary, adopts cascade images as input. Only low-resolution input with rather limited computation goes through the main segmentation branch. The new cascade feature fusion unit and cascade label guidance can successfully utilize higher-resolution input to recover and refine the low-resolution branch prediction. It is a high-efficiency system with good-quality segmentation results.

\section{Image Cascade Network}
\begin{figure}[t]
	\centering
	\includegraphics[width=0.98\linewidth]{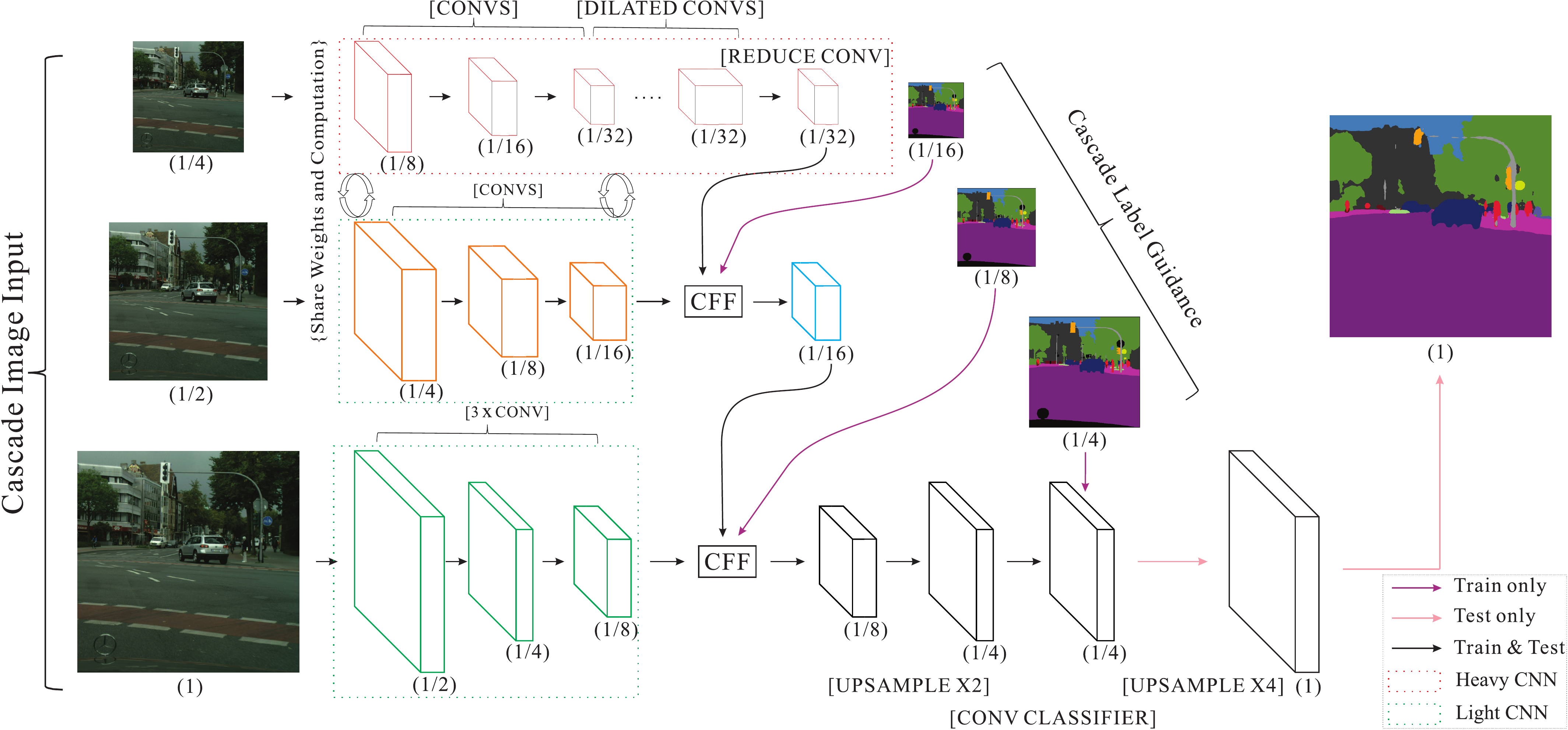}
	\caption{Network architecture of ICNet. `CFF' stands for cascade feature fusion detailed in Sec.~\ref{sec:cff}. Numbers in parentheses are feature map size ratios to the full-resolution input. Operations are highlighted in brackets. The final $\times 4$ upsampling in the bottom branch is only used during testing.}
	\label{fig:imagecascadenetwork}
\end{figure}

We start by analyzing computation time budget of different components on the high performance segmentation framework PSPNet~\cite{zhao2017pspnet} with experimental statistics. Then we introduce the \textit{image cascade network} (ICNet) as illustrated in Fig.~\ref{fig:imagecascadenetwork}, along with the cascade feature fusion unit and cascade label guidance, for fast semantic segmentation.

\subsection{Speed Analysis}\label{sec:speed}
In convolution, the transformation function $\varPhi$ is applied to input feature map $V \in \mathbb{R}^{c \times h \times w}$ to obtain the output map $U \in \mathbb{R}^{c' \times h' \times w'}$, where $c$, $h$ and $w$ denote features channel, height and width respectively.
The transformation operation $\varPhi: V \rightarrow U$ is achieved by applying $c'$ number of 3D kernels $K \in \mathbb{R}^{c \times k \times k}$ where $k \times k$ (e.g, $3 \times 3$) is kernel spatial size.
Thus the total number of operations $O(\varPhi)$ in convolution layer is $c'ck^{2}h'w'$. The spatial size of the output map $h'$ and $w'$ are highly related to the input, controlled by parameter stride $s$ as $h'=h/s, w'=w/s$, making 
\begin{equation}\label{eq:time}
	O(\varPhi) \approx c'ck^{2}hw/s^{2}.
\end{equation}
The computation complexity is associated with feature map resolution (e.g., $h$, $w$, $s$), number of kernels and network width (e.g., $c$, $c'$). Fig.~\ref{fig:accuracytime-timeimageresolution}(b) shows the time cost of two resolution images in PSPNet50. Blue curve corresponds to high-resolution input with size $1024 \times 2048$ and green curve is for image with resolution $512 \times 1024$. Computation increases squarely regarding image resolution. For either curve, feature maps in stage4 and stage5 are with the same spatial resolution, i.e., $1/8$ of the original input; but the computation in stage5 is four times heavier than that in stage4. It is because convolutional layers in stage5 double the number of kernels $c$ together with input channel $c'$.

\subsection{Network Architecture}
According to above time budget analysis, we adopt intuitive speedup strategies in experiments to be detailed in Sec.~\ref{sec:experiment}, including downsampling input, shrinking feature maps and conducting model compression. The corresponding results show that it is very difficult to keep a good balance between inference accuracy and speed. The intuitive strategies are effective to reduce running time, while they yield very coarse prediction maps. Directly feeding high-resolution images into a network is unbearable in computation.

Our proposed system \textit{image cascade network} (ICNet) does not simply choose either way. Instead it takes cascade image inputs (i.e., low-, medium- and high resolution images), adopts cascade feature fusion unit (Sec.~\ref{sec:cff}) and is trained with cascade label guidance (Sec.~\ref{sec:clg}). The new architecture is illustrated in Fig.~\ref{fig:imagecascadenetwork}. 
The input image with full resolution (e.g., $1024 \times 2048$ in Cityscapes~\cite{cordts2016cityscapes}) is downsampled by factors of $2$ and $4$, forming cascade input to medium- and high-resolution branches.

Segmenting the high-resolution input with classical frameworks like FCN directly is time consuming. To overcome this shortcoming, we get semantic extraction using low-resolution input as shown in top branch of Fig.~\ref{fig:imagecascadenetwork}. A $1/4$ sized image is fed into PSPNet with downsampling rate 8, resulting in a $1/32$-resolution feature map. To get high quality segmentation, medium and high resolution branches (middle and bottom parts in Fig.~\ref{fig:imagecascadenetwork}) help recover and refine the coarse prediction. Though some details are missing and blurry boundaries are generated in the top branch, it already harvests most semantic parts. Thus we can safely limit the number of parameters in both middle and bottom branches. Light weighted CNNs (green dotted box) are adopted in higher resolution branches; different-branch output feature maps are fused by cascade-feature-fusion unit (Sec.~\ref{sec:cff}) and trained with cascade label guidance (Sec.~\ref{sec:clg}).

Although the top branch is based on a full segmentation backbone, the input resolution is low, resulting in limited computation. Even for PSPNet with 50+ layers, inference time and memory are 18ms and 0.6GB for the large images in Cityscapes. Because weights and computation (in 17 layers) can be shared between low- and medium-branches, only 6ms is spent to construct the fusion map. Bottom branch has even less layers. Although the resolution is high, inference only takes 9ms. Details of the architecture are presented in the supplementary file. With all these three branches, our ICNet becomes a very efficient and memory friendly architecture that can achieve good-quality segmentation.

\subsection{Cascade Feature Fusion}\label{sec:cff}
\begin{wrapfigure}{rt}{0.4\textwidth}
	\includegraphics[width=0.4\textwidth]{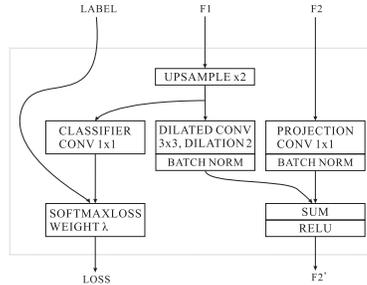}
	\caption{Cascade feature fusion.}% Given two input feature maps $F_1$ and $F_2$ along with ground truth `LABEL', the spatial resolution of $F_2$ is twice of $F_1$ and the output feature map $F_2'$ is with the same spatial size as $F_2$.}
	\label{fig:cascadefeaturefusion}
\end{wrapfigure}

To combine cascade features from different-resolution inputs, we propose a cascade feature fusion (CFF) unit as shown in Fig.~\ref{fig:cascadefeaturefusion}. The input to this unit contains three components: two feature maps $F_1$ and $F_2$ with sizes $C_1 \times H_1 \times W_1$ and $C_2 \times H_2 \times W_2$ respectively, and a ground-truth label with resolution $1 \times H_2 \times W_2$. $F_2$ is with doubled spatial size of $F_1$.

We first apply upsampling rate 2 on $F_1$ through bilinear interpolation, yielding the same spatial size as $F_2$. Then a dilated convolution layer with kernel size $C_3 \times 3 \times 3$ and dilation 2 is applied to refine the upsampled features. The resulting feature is with size $C_3 \times H_2 \times W_2$. This dilated convolution combines feature information from several originally neighboring pixels. Compared with deconvolution, upsampling followed by dilated convolution only needs small kernels, to harvest the same receptive field. To keep the same receptive field, deconvolution needs larger kernel sizes than upsampling with dilated convolution (i.e., $7 \times 7$ vs. $3 \times 3$), which causes more computation.

For feature $F_2$, a projection convolution with kernel size $C_3 \times 1 \times 1$ is utilized to project $F_2$ so that it has the same number of channels as the output of $F_1$. Then two batch normalization layers are used to normalize these two processed features as shown in Fig.~\ref{fig:cascadefeaturefusion}. Followed by an element-wise `sum' layer and a `ReLU' layer, we obtain the fused feature
$F_2'$ as $C_3 \times H_2 \times W_2$. To enhance learning of $F_1$, we use an auxiliary label guidance on the upsampled feature of $F_1$.

\subsection{Cascade Label Guidance}\label{sec:clg}
To enhance the learning procedure in each branch, we adopt a cascade label guidance strategy. It utilizes different-scale (e.g., $1/16$, $1/8$, and $1/4$) ground-truth labels to guide the learning stage of low, medium and high resolution input. Given $\mathcal{T}$ branches (i.e., $\mathcal{T}$=3) and $\mathcal{N}$ categories. In branch $t$, the predicted feature map $\mathcal{F}^t$ has spatial size $\mathcal{Y}_t \times \mathcal{X}_t$. The value at position $(n,y,x)$ is $\mathcal{F}^{t}_{n,y,x}$. The corresponding ground truth label for 2D position $(y,x)$ is $\hat{n}$. To train ICNet, we append weighted softmax cross entropy loss in each branch with related loss weight $\lambda_t$. Thus we minimize the loss function $\mathcal{L}$ defined as
\begin{equation}\label{eq:loss}
	\mathcal{L} = - \sum_{t=1}^{\mathcal{T}}\lambda_t\frac{1}{\mathcal{Y}_t \mathcal{X}_t}\sum_{y=1}^{\mathcal{Y}_t}\sum_{x=1}^{\mathcal{X}_t}\log{\frac{e^{\mathcal{F}^{t}_{\hat{n},y,x}}}{\sum_{n=1}^{\mathcal{N}} e^{\mathcal{F}^{t}_{n,y,x}}}}.
\end{equation}
In the testing phase, the low and medium guidance operations are simply abandoned, where only high-resolution branch is retained. 
This strategy makes gradient optimization smoother for easy training. With more powerful learning ability in each branch, the final prediction map is not dominated by any single branch. 

%------------------------------------------------------------------------
\section{Structure Comparison and Analysis}
\begin{figure}[t]
	\begin{center}
		\includegraphics[width=1.0\linewidth]{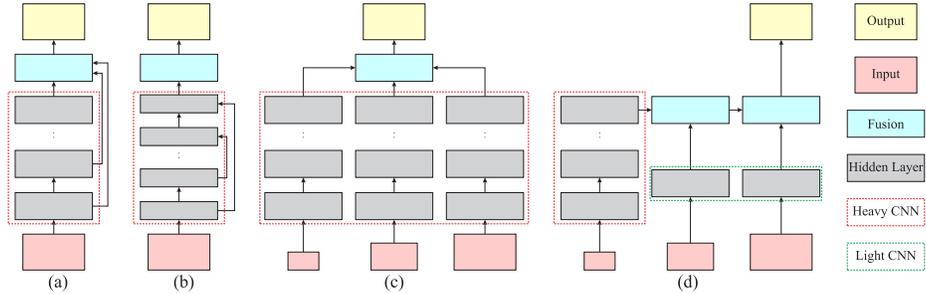}
	\end{center}
	\caption{Comparison of semantic segmentation frameworks. (a) Intermediate skip connection used by FCN~\cite{long2015fully} and Hypercolumns~\cite{hariharan2015hypercolumns}. (b) Encoder-decoder structure incorporated in SegNet~\cite{badrinarayanan2015segnet}, DeconvNet~\cite{noh2015learning}, UNet~\cite{ronneberger2015unet}, ENet~\cite{paszke2016enet}, and step-wise reconstruction \& refinement from LRR~\cite{ghiasi2016laplacian} and RefineNet~\cite{lin2017refine}. (c) Multi-scale prediction ensemble adopted by DeepLab-MSC~\cite{chen2015semantic} and PSPNet-MSC~\cite{zhao2017pspnet}. (d) Our ICNet architecture.}
	\label{fig:diffstructure}
\end{figure}

Now we illustrate the difference of ICNet from existing cascade architectures for semantic segmentation. Typical structures in previous semantic segmentation systems are illustrated in Fig.~\ref{fig:diffstructure}. Our proposed ICNet (Fig.~\ref{fig:diffstructure}(d)) is by nature different from others. Previous frameworks are all with relatively intensive computation given the high-resolution input. While in our cascade structure, only the lowest-resolution input is fed into the heavy CNN with much reduced computation to get the coarse semantic prediction. The higher-res inputs are designed to recover and refine the prediction progressively regarding blurred boundaries and missing details. Thus they are processed by light-weighted CNNs. Newly introduced cascade-feature-fusion unit and cascade label guidance strategy integrate medium and high resolution features to refine the coarse semantic map gradually. In this special design, ICNet achieves high-efficiency inference with reasonable-quality segmentation results. 

%-------------------------------------------------------------------------
\section{Experimental Evaluation}\label{sec:experiment}
Our method is effective for high resolution images. We evaluate the architecture on three challenging datasets, including urban-scene understanding dataset Cityscapes~\cite{cordts2016cityscapes} with image resolution $1024 \times 2048$, CamVid~\cite{BrostowFC09} with image resolution $720 \times 960$ and stuff understanding dataset COCO-Stuff~\cite{caesar2016coco} with image resolution up to $640 \times 640$. There is a notable difference between COCO-Stuff and object/scene segmentation datasets of VOC2012~\cite{everingham2010pascal} and ADE20K~\cite{zhou2016semantic}. In the latter two sets, most images are of low resolution (e.g., $300 \times 500$), which can already be processed quickly. While in COCO-Stuff, most images are larger, making it more difficult to achieve real-time performance.

In the following, we first show intuitive speedup strategies and their drawbacks, then reveal our improvement with quantitative and visual analysis. 

\subsection{Implementation Details}
We conduct experiments based on platform Caffe~\cite{jia2014caffe}. All experiments are on a workstation with Maxwell TitanX GPU cards under CUDA 7.5 and CUDNN V5. Our testing uses only one card. To measure the forward inference time, we use the time measure tool `Caffe time' and set the repeating iteration number to 100 to eliminate accidental errors during testing. All the parameters in batch normalization layers are merged into the neighboring front convolution layers. 

For the training hyper-parameters, the mini-batch size is set to 16. The base learning rate is 0.01 and the `poly' learning rate policy is adopted with power 0.9, together with the maximum iteration number set to 30K for Cityscapes, 10K for CamVid and 30K for COCO-Stuff. Momentum is 0.9 and weight decay is 0.0001. Data augmentation contains random mirror and rand resizing between 0.5 and 2. The auxiliary loss weights are empirically set to 0.4 for $\lambda_1$ and $\lambda_2$, 1 for $\lambda_3$ in Eq.~\ref{eq:loss}, as adopted in~\cite{zhao2017pspnet}. For evaluation, both \textit{mean of class-wise intersection over union} (mIoU) and \textit{network forward time} (Time) are used.

\subsection{Cityscapes}
We first apply our framework to the recent urban scene understanding dataset Cityscapes~\cite{cordts2016cityscapes}. This dataset contains high-resolution $1024 \times 2048$ images, which make it a big challenge for fast semantic segmentation. It contains 5,000 finely annotated images split into training, validation and testing sets with 2,975, 500, and 1,525 images respectively. The dense annotation contains 30 common classes of road, person, car, etc. 19 of them are used in training and testing.

\subsubsection{Intuitive Speedup}
According to the time complexity shown in Eq.~\eqref{eq:time}, we do intuitive speedup in three aspects, namely downsampling input, downsampling feature, and model compression.

\begin{figure}[bpt]
	\centering
	\begin{tabular}{ccc}
		\includegraphics[width=0.32\linewidth]{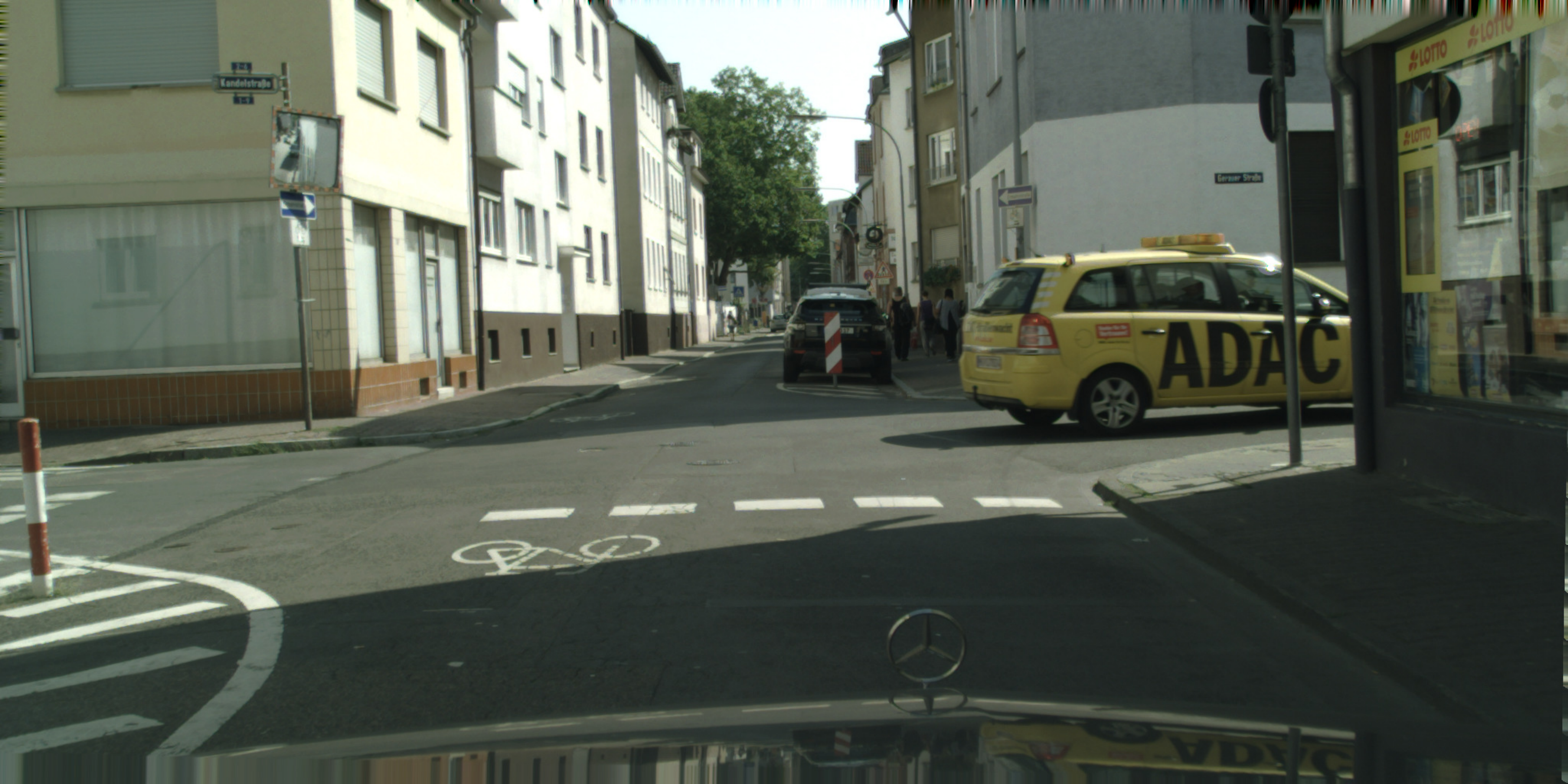}&
		\includegraphics[width=0.32\linewidth]{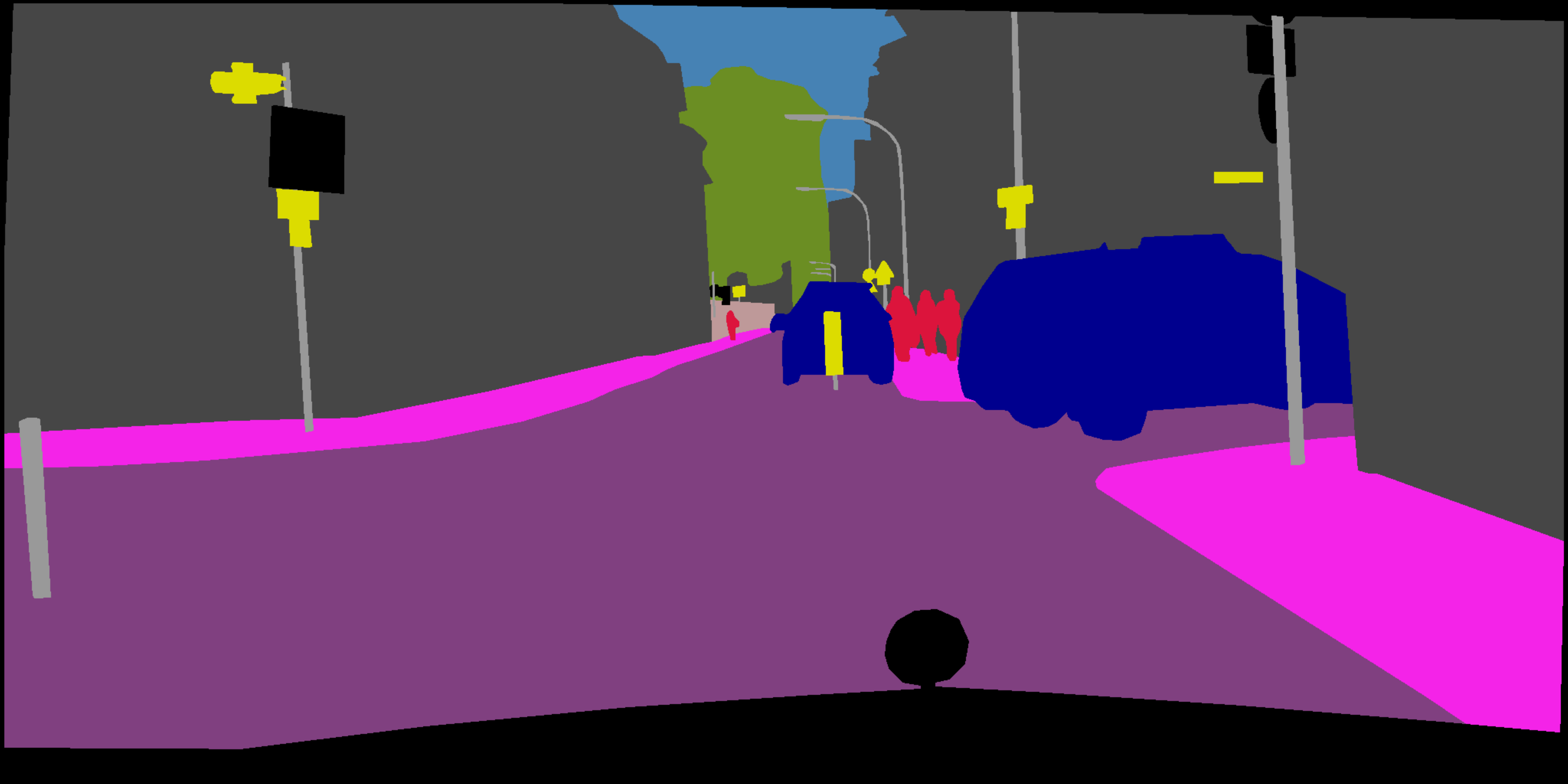}&
		\includegraphics[width=0.32\linewidth]{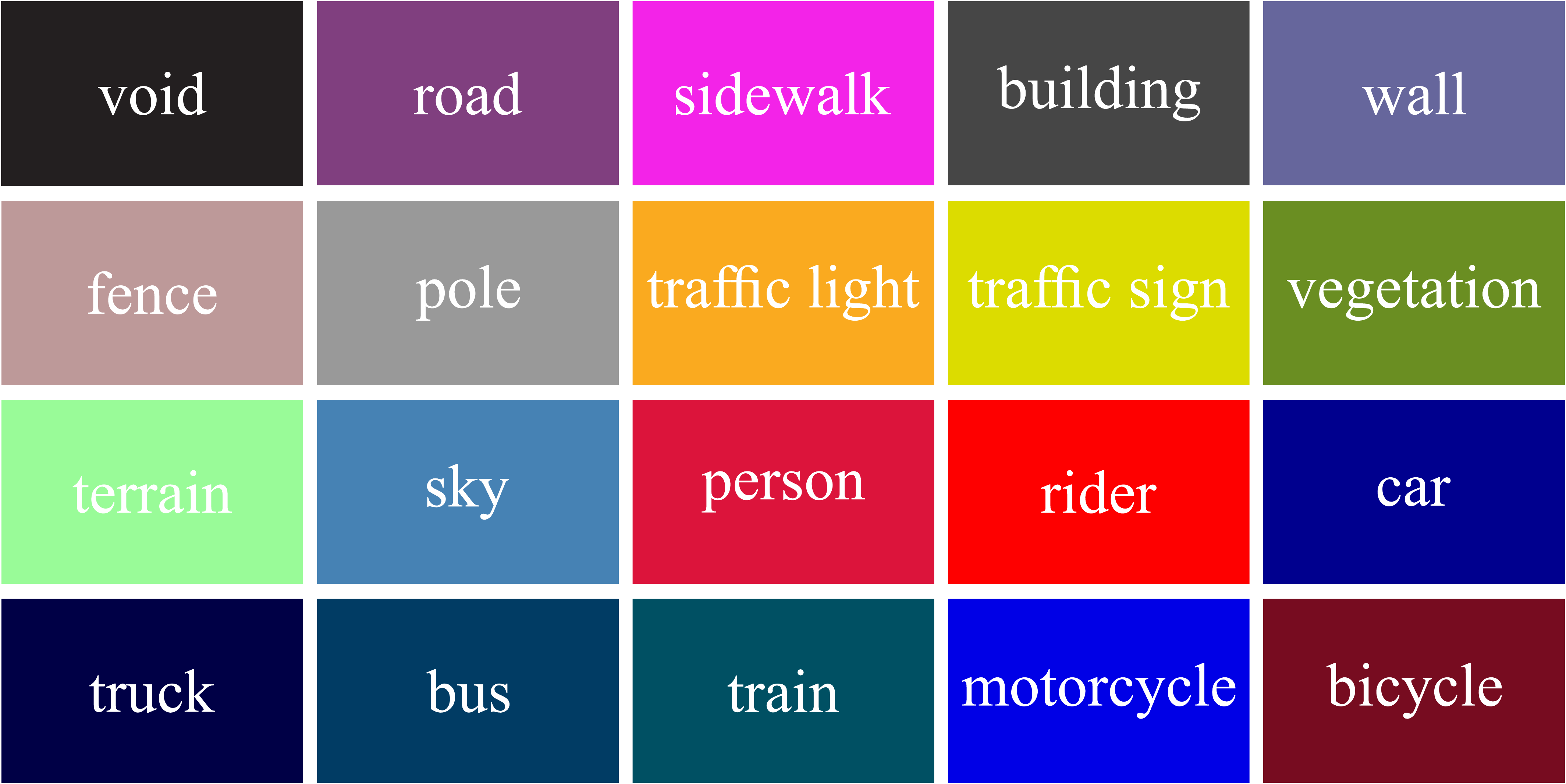}\\
		{\scriptsize (a) input image} & {\scriptsize (b) ground truth} & {\scriptsize (c) colormap}\\
		\includegraphics[width=0.32\linewidth]{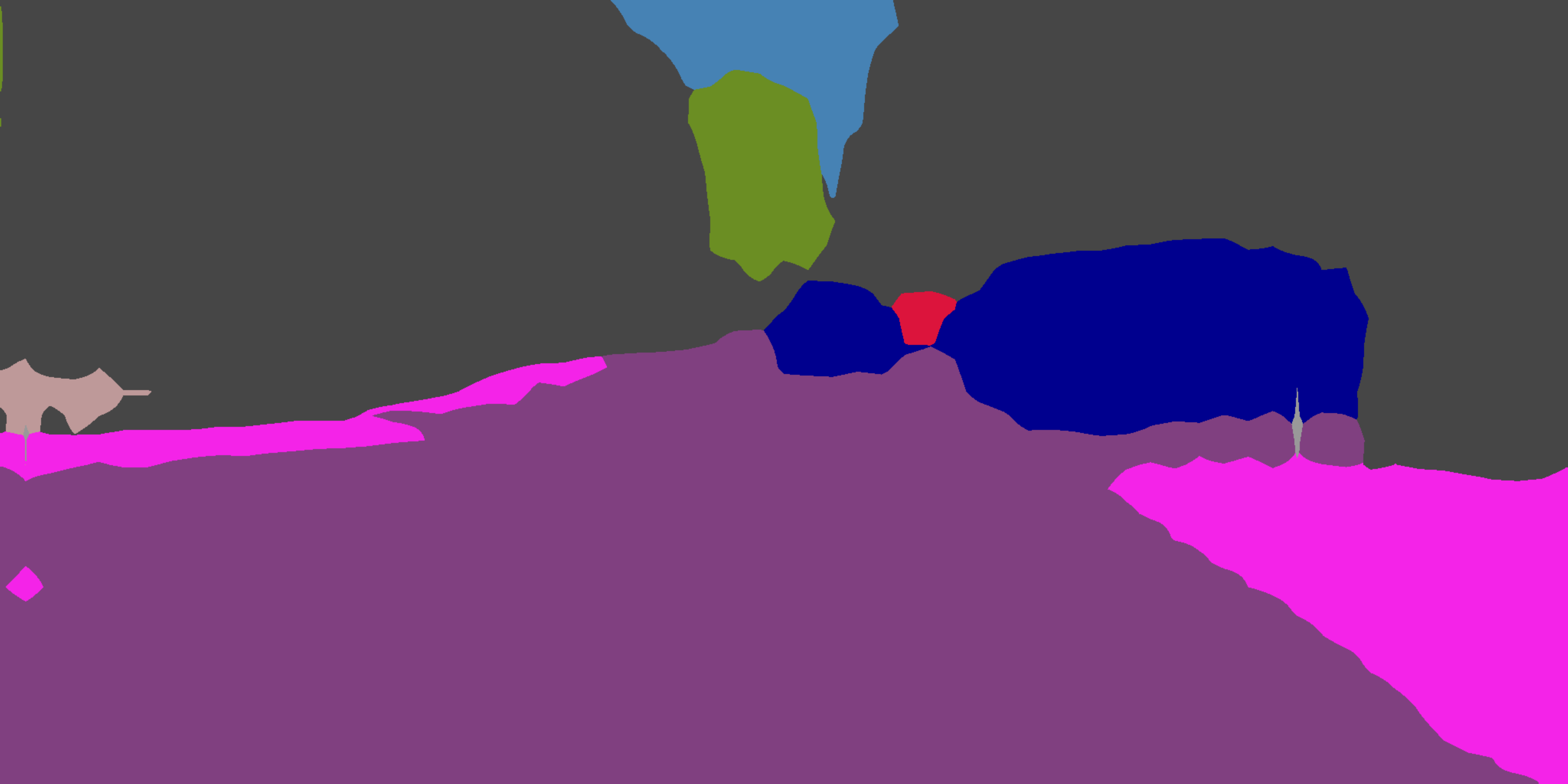}&
		\includegraphics[width=0.32\linewidth]{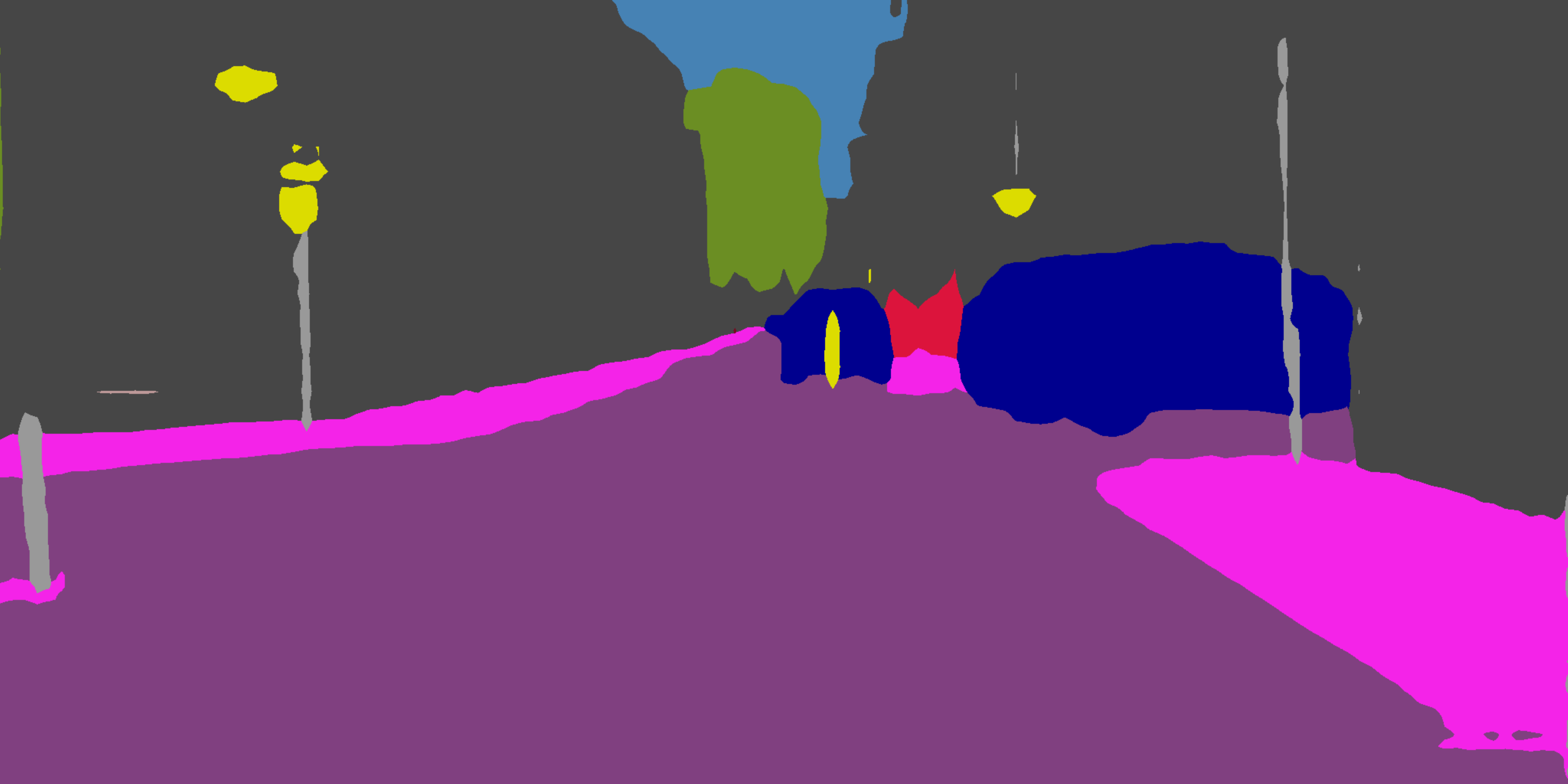}&
		\includegraphics[width=0.32\linewidth]{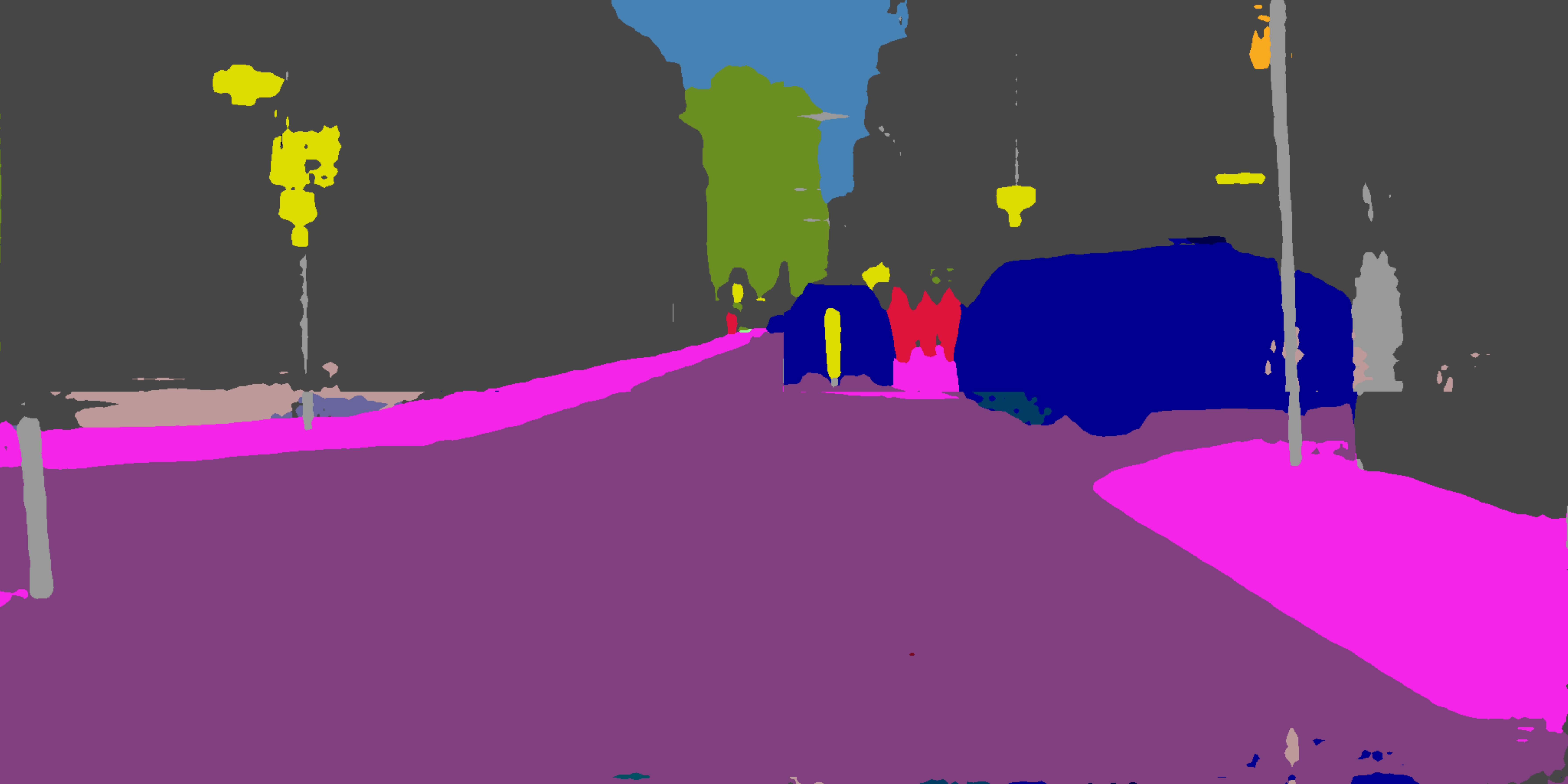}\\
		{\scriptsize (d) scale 0.25 (42ms/60.7\%)} & {\scriptsize (e) scale 0.5 (123ms/68.4\%)} & {\scriptsize (f) scale 1 (446ms/71.7\%)}\\
	\end{tabular}
	\caption{Downsampling input: prediction of PSPNet50 on the validation set of Cityscapes. Values in the parentheses are the inference time and mIoU.}
	\label{fig:resolutionprediction}
\end{figure}

\begin{table}[bpt]
	\caption{\textbf{Left:} Downsampling feature with factors 8, 16 and 32. \textbf{Right:} Model compression with kernel keeping rates 1, 0.5 and 0.25.}
	\label{tab:timedownsamplesize-timemodelcompression}
	\begin{minipage}[t]{0.48\linewidth}
		\centering
		\setlength{\tabcolsep}{3pt}
		\begin{tabular}{l | c c c}
			\toprule[1pt]
			Downsample Size & 8 & 16 & 32 \\
			\hline
			\hline
			mIoU (\%) & 71.7 & 70.2 & 67.1 \\
			Time (ms) & 446 & 177 & 131 \\
			\bottomrule[1pt]
		\end{tabular}
	\end{minipage}
	\hfill
	\begin{minipage}[t]{0.48\linewidth}
		\centering
		\setlength{\tabcolsep}{3pt}
		\begin{tabular}{l | c c c}
			\toprule[1pt]
			Kernel Keeping Rates & 1 & 0.5 & 0.25 \\
			\hline
			\hline
			mIoU (\%) & 71.7 & 67.9 & 59.4 \\
			Time (ms) & 446 & 170 & 72 \\
			\bottomrule[1pt]
		\end{tabular}
	\end{minipage}
\end{table}

\paragraph{Downsampling Input}
Image resolution is the most critical factor that affects running speed as analyzed in Sec.~\ref{sec:speed}. A simple approach is to use the small-resolution image as input. We test downsampling the image with ratios $1/2$ and $1/4$, and feeding the resulting images into PSPNet50. We directly upsample prediction results to the original size. This approach empirically has several drawbacks as illustrated in Fig.~\ref{fig:resolutionprediction}. With scaling ratio 0.25, although the inference time is reduced by a large margin, the prediction map is very coarse, missing many small but important details compared to the higher resolution prediction. With scaling ratio 0.5, the prediction recovers more information compared to the 0.25 case. Unfortunately, the person and traffic light far from the camera are still missing and object boundaries are blurred. To make things worse, the running time is still too long for a real-time system.

\paragraph{Downsampling Feature}
Besides directly downsampling the input image, another simple choice is to scale down the feature map by a large ratio in the inference process. FCN~\cite{long2015fully} downsampled it for 32 times and DeepLab~\cite{chen2015semantic} did that for 8 times. We test PSPNet50 with downsampling ratios of 1:8, 1:16 and 1:32 and show results in the left of Table~\ref{tab:timedownsamplesize-timemodelcompression}. A smaller feature map can yield faster inference at the cost of sacrificing prediction accuracy. The lost information is mostly detail contained in low-level layers. Also, even with the smallest resulting feature map under ratio 1:32, the system still takes 131ms in inference. 

\paragraph{Model Compression}
Apart from the above two strategies, another natural way to reduce network complexity is to trim kernels in each layer. Compressing models becomes an active research topic in recent years due to the high demand. The solutions \cite{Iandola2016squeezenet,han2016deepcom,han2017dsd,li2017pruning} can make a complicated network reduce to a lighter one under user-controlled accuracy reduction. We adopt recent effective classification model compression strategy presented in \cite{li2017pruning} on our segmentation models. For each filter, we first calculate the sum of kernel $\ell$$_1$-norm. Then we sort these sum results in a descending order and keep only the most significant ones. Disappointingly, this strategy also does not meet our requirement given the compressed models listed in the right of Table~\ref{tab:timedownsamplesize-timemodelcompression}. Even by keeping only a quarter of kernels, the inference time is still too long. Meanwhile the corresponding mIoU is intolerably low -- it already cannot produce reasonable segmentation for many applications.

\begin{table}[bpt]
	\setlength{\tabcolsep}{6pt}
	\caption{Performance of ICNet with different branches on validation set of Citysapes. The baseline method is PSPNet50 compressed to a half. `sub4', `sub24' and `sub124' represent predictions in low-, medium-, and high-resolution branches respectively.}
	\label{tab:imagecascadeperformance}
	\begin{center}
		\begin{tabular}{l | c c c c}
			\toprule[1pt]
			Items & Baseline & sub4 & sub24 & sub124 \\
			\hline
			\hline
			mIoU (\%) & 67.9 & 59.6 & 66.5 & \textbf{67.7} \\
			Time (ms) & 170 & 18 & 25 & 33 \\
			Frame (fps) & 5.9 & 55.6 & 40 & \textbf{30.3} \\
			Speedup & 1$\times$ & 9.4$\times$ & 6.8$\times$ & \textbf{5.2}$\times$ \\
			Memory (GB) & 9.2 & 0.6 & 1.1 & 1.6 \\
			Memory Save & 1$\times$ & 15.3$\times$ & 8.4$\times$ & \textbf{5.8}$\times$\\
			\bottomrule[1pt]
		\end{tabular}
	\end{center}
\end{table}

\begin{table}[bpt]
	\setlength{\tabcolsep}{4pt}
	\caption{Effectiveness of cascade feature fusion unit (CFF) and cascade label guidance (CLG). `DC3', `DC5' and `DC7' denote replacing `bilinear upsampling + dilated convolution' with deconvolution operation with kernels $3 \times 3$, $5 \times 5$ and $7 \times 7$ respectively.}
	\label{tab:cff_clg}
	\begin{center}
		\begin{tabular}{c c c c c | c c}
			\toprule[1pt]
			DC3 & DC5 & DC7 & CFF & CLG & mIoU (\%) & Time (ms) \\
			\hline
			\hline
			\checkmark & & & & \checkmark & 66.7 & 31 \\
			& \checkmark & & & \checkmark & 66.7 & 34 \\
			& & \checkmark & & \checkmark & 68.0 & 38 \\
			\rowcolor{Gray}
			& & & \checkmark & \checkmark & 67.7 & 33 \\
			& & & \checkmark & & 66.8 & 33 \\
			\bottomrule[1pt]
		\end{tabular}
	\end{center}
\end{table}

\subsubsection{Cascade Branches}
We do ablation study on cascade branches, the results are shown in Table~\ref{tab:imagecascadeperformance}. Our baseline is the half-compressed PSPNet50, 170ms inference time is yielded with mIoU reducing to 67.9\%. They indicate that model compression has almost no chance to achieve real-time performance under the condition of keeping decent segmentation quality. Based on this baseline, we test our ICNet on different branches. To show the effectiveness of the proposed cascade framework, we denote the outputs of low-, medium- and high-resolution branches as `sub4', `sub24' and `sub124', where the numbers stand for the information used. The setting `sub4' only uses the top branch with the low-resolution input. `sub24' and `sub124' respectively contain top two and all three branches.

We test these three settings on the validation set of Cityscapes and list the results in Table~\ref{tab:imagecascadeperformance}. With just the low-resolution input branch, although running time is short, the result quality drops to 59.6\%. Using two and three branches, we increase mIoU to 66.5\% and 67.7\% respectively. The running time only increases by 7ms and 8ms. Note our segmentation quality nearly stays the same as the baseline, and yet is 5.2$\times$ times faster. The memory consumption is significantly reduced by 5.8$\times$.

\subsubsection{Cascade Structure}
We also do ablation study on cascade feature fusion unit and cascade label guidance. The results are shown in Table~\ref{tab:cff_clg}. Compared to the deconvolution layer with $3 \times 3$ and $5 \times 5$ kernels, with similar inference efficiency, cascade feature fusion unit gets higher mIoU performance. Compared to deconvolution layer with a larger kernel with size $7 \times 7$, the mIoU performance is close, while cascade feature fusion unit yields faster processing speed. Without the cascade label guidance, the performance drops a lot as shown in the last row.

\begin{table}[bpt]
	\setlength{\tabcolsep}{3pt}
	\caption{Predicted mIoU and inference time on Cityscapes test set with image resolution $1024 \times 2048$. `DR' stands for image downsampling ratio during testing (e.g, DR=4 represents testing at resolution $256 \times 512$). Methods trained using both fine and coarse data are marked with `$\dag$'.}
	\label{tab:cstestperformance}
	\begin{center}
		\begin{tabular}{l | c c c c}
			\toprule[1pt]
			Method & DR & mIoU (\%) & Time (ms) & Frame (fps) \\
			\hline
			\hline
			SegNet~\cite{badrinarayanan2015segnet} & 4 & 57.0 & 60 & 16.7 \\
			ENet~\cite{paszke2016enet} & 2 & 58.3 & 13 & 76.9 \\
			SQ~\cite{treml2016speeding} & no & 59.8 & 60 & 16.7 \\
			CRF-RNN~\cite{zheng2015conditional} & 2 & 62.5 & 700 & 1.4 \\
			DeepLab~\cite{chen2015semantic} & 2 & 63.1 & 4000 & 0.25 \\
			FCN-8S~\cite{long2015fully} & no & 65.3 & 500 & 2 \\
			Dilation10~\cite{yu2016multi} & no & 67.1 & 4000 & 0.25 \\
			FRRN~\cite{pohlen2017FRRN} & 2 & 71.8 & 469 & 2.1 \\
			PSPNet\footnotemark~\cite{zhao2017pspnet} & no & 81.2 & 1288 & 0.78 \\
			%DeepLabv2~\cite{chen2016deeplab} & no & 70.4 & 
			%\hline
			%\hline
			%Adelaide~\cite{lin2015efficient} & no & 71.6 & 35000 & 0.03 \\
			%ResNet38~\cite{wu2016wider} & no & 80.6 & 3000 & 0.33 \\
			%PSPNet~\cite{zhao2017pspnet}
			%\hline
			%RefineNet~\cite{lin16refine} & no & 73.6 & 1174 & 0.9 \\
			\hline
			\hline
			\rowcolor{Gray}
			ICNet & no & 69.5 & 33 & 30.3 \\
			\rowcolor{Gray}
			ICNet$^\dag$ & no & 70.6 & 33 & 30.3 \\
			\bottomrule[1pt]
		\end{tabular}
	\end{center}
\end{table}
\footnotetext{Single network forward costs 1288ms (with TitanX Maxwell, 680ms for Pascal) while mIoU aimed testing for boosting performance (81.2\% mIoU) costs 51.0s.}
%with code \href{https://github.com/hszhao/PSPNet}{https://github.com/hszhao/PSPNet}.}

\subsubsection{Methods Comparison}
We finally list mIoU performance and inference time of our proposed ICNet on the test set of Cityscapes. It is trained on training and validation sets of Cityscapes for 90K iterations. Results are included in Table~\ref{tab:cstestperformance}. The reported mIoUs and running time of other methods are shown in the official Cityscapes leadboard. For fairness, we do not include methods without reporting running time. Many of these methods may have adopted time-consuming multi-scale testing for the best result quality.

Our ICNet yields mIoU 69.5\%. It is even quantitatively better than several methods that do not care about speed. It is about 10 points higher than ENet \cite{paszke2016enet} and SQ~\cite{treml2016speeding}. Training with both fine and coarse data boosts mIoU performance to 70.6\%.  ICNet is a 30fps method on $1024 \times 2048$ resolution images using only one TitanX GPU card. Video example can be accessed through link\footnote{\href{https://youtu.be/qWl9idsCuLQ}{https://youtu.be/qWl9idsCuLQ}}.

\begin{figure}[bpt]
	\centering
	\begin{tabular}{ccc}
		\includegraphics[width=0.32\linewidth]{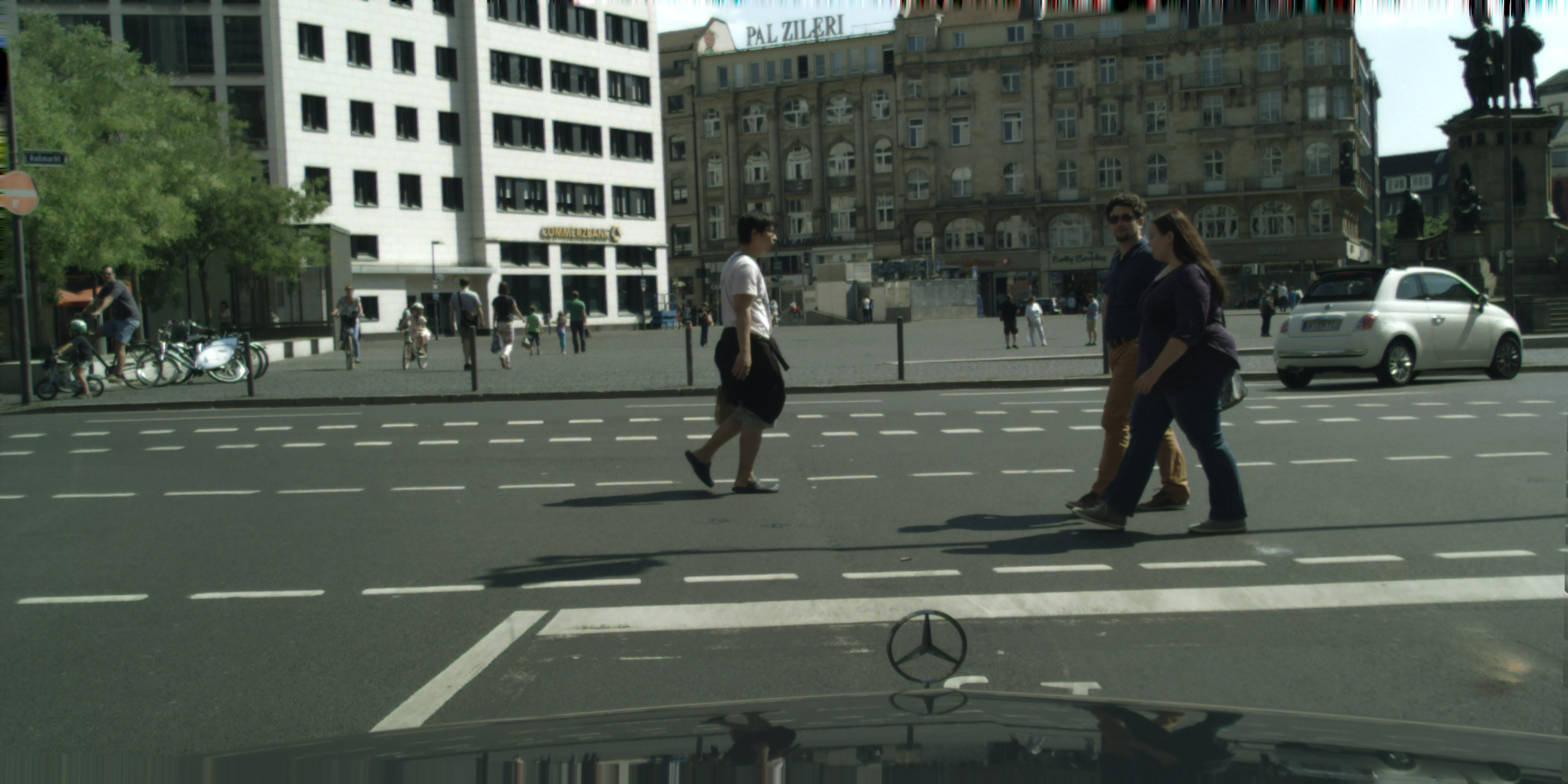}&
		\includegraphics[width=0.32\linewidth]{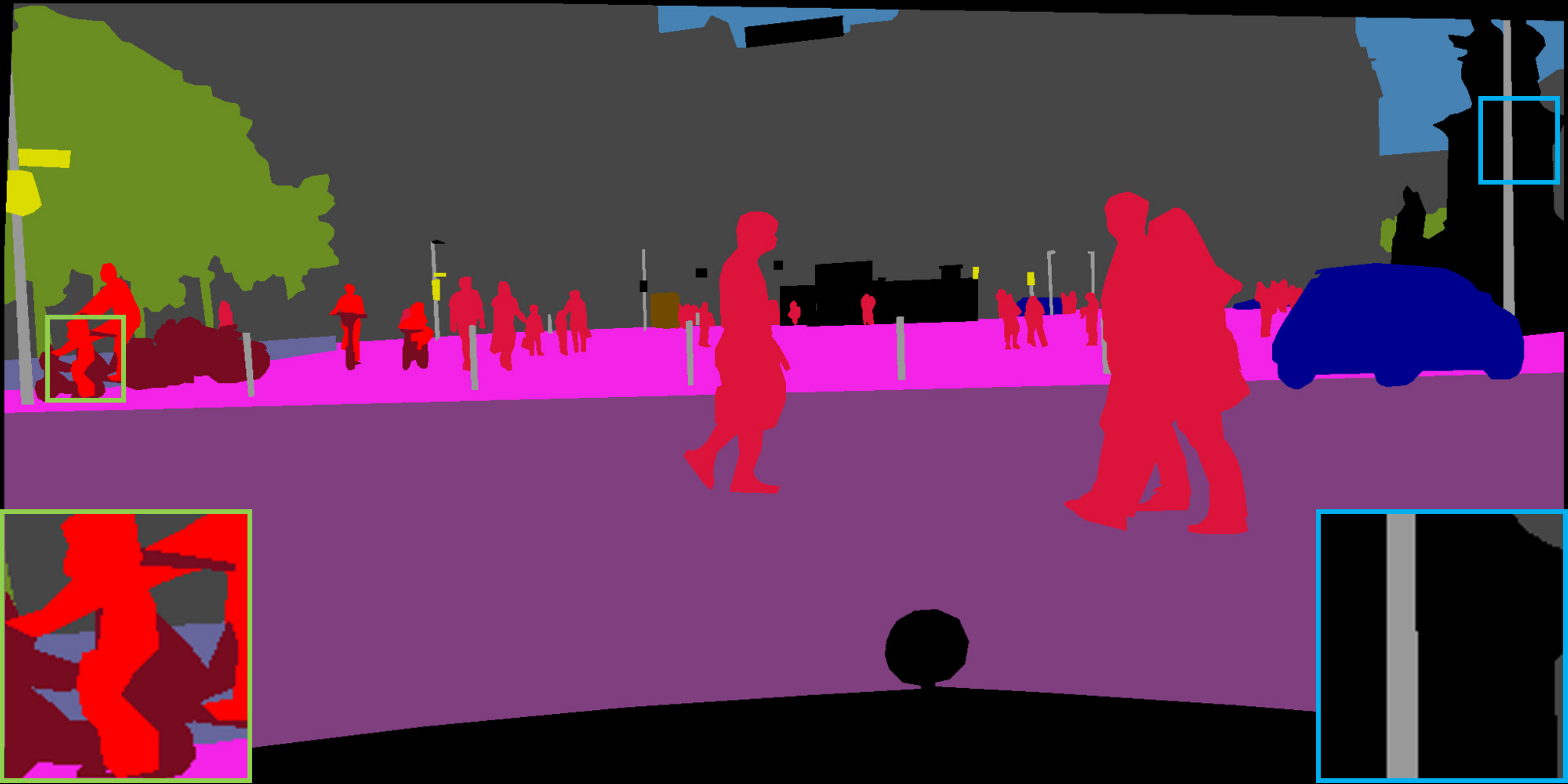}&
		\includegraphics[width=0.32\linewidth]{images/color.pdf}\\
		{\scriptsize (a) input image} & {\scriptsize (b) ground truth} & {\scriptsize (c) colormap}\\
		\includegraphics[width=0.32\linewidth]{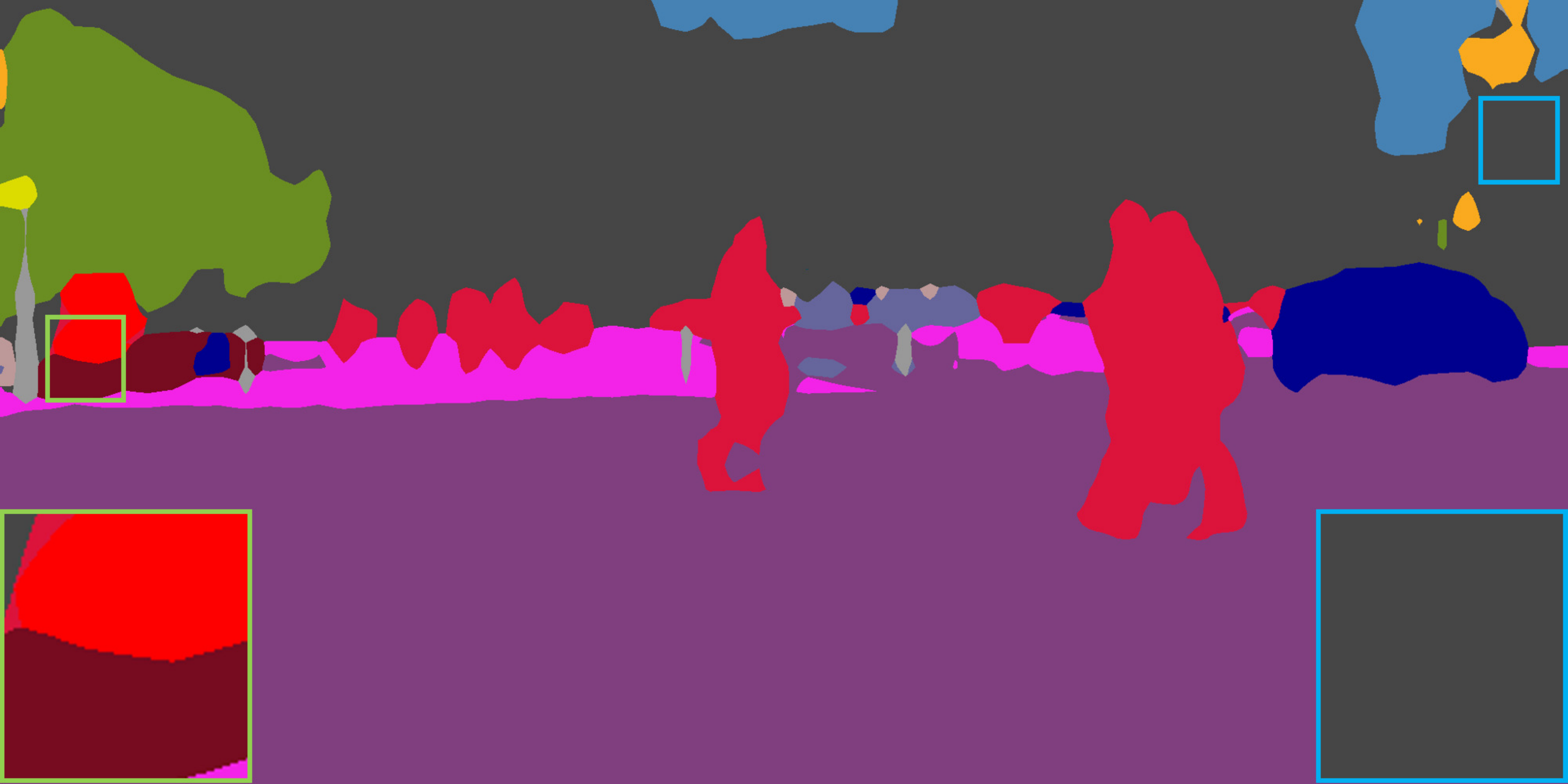}&
		\includegraphics[width=0.32\linewidth]{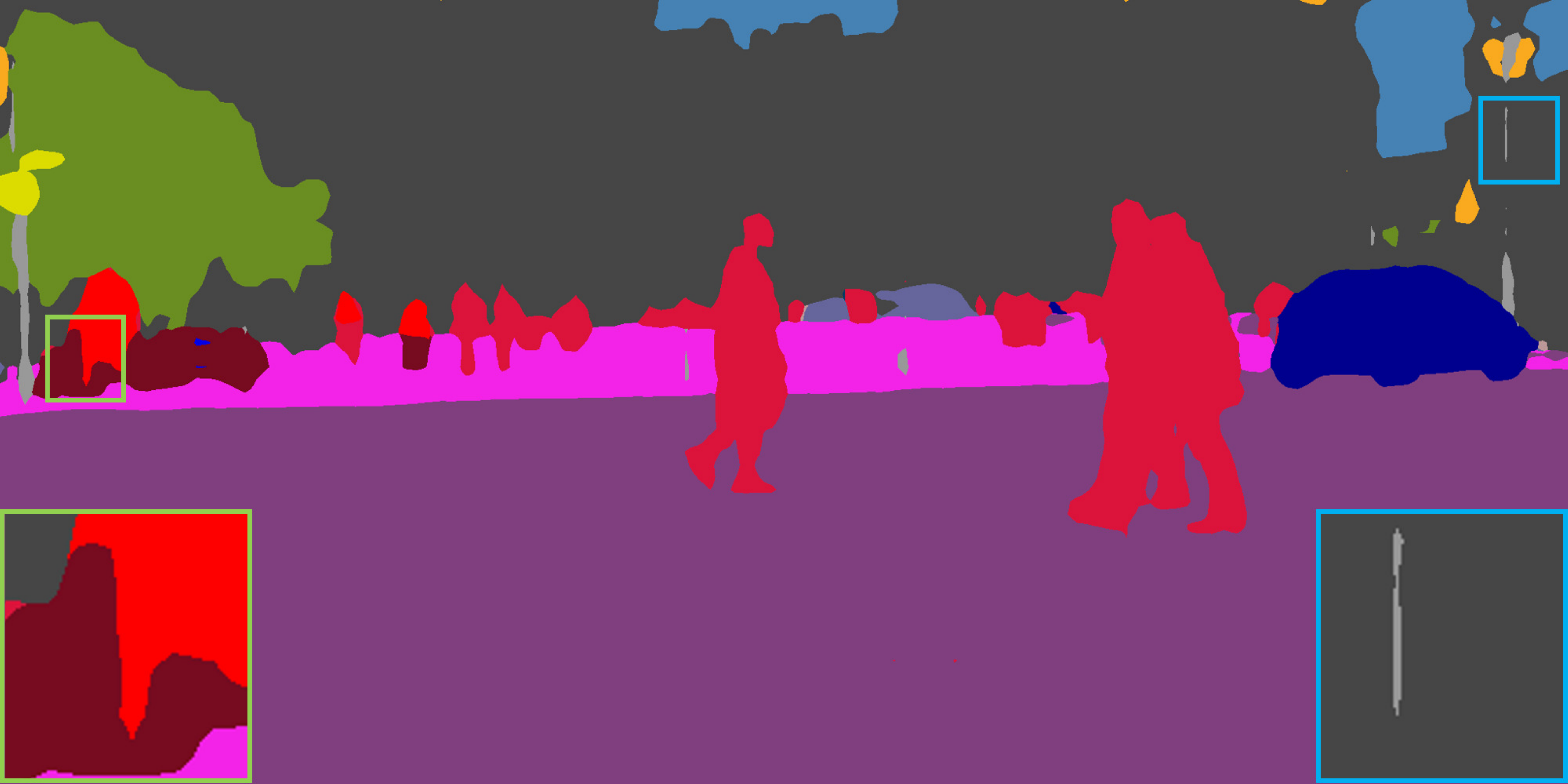}&
		\includegraphics[width=0.32\linewidth]{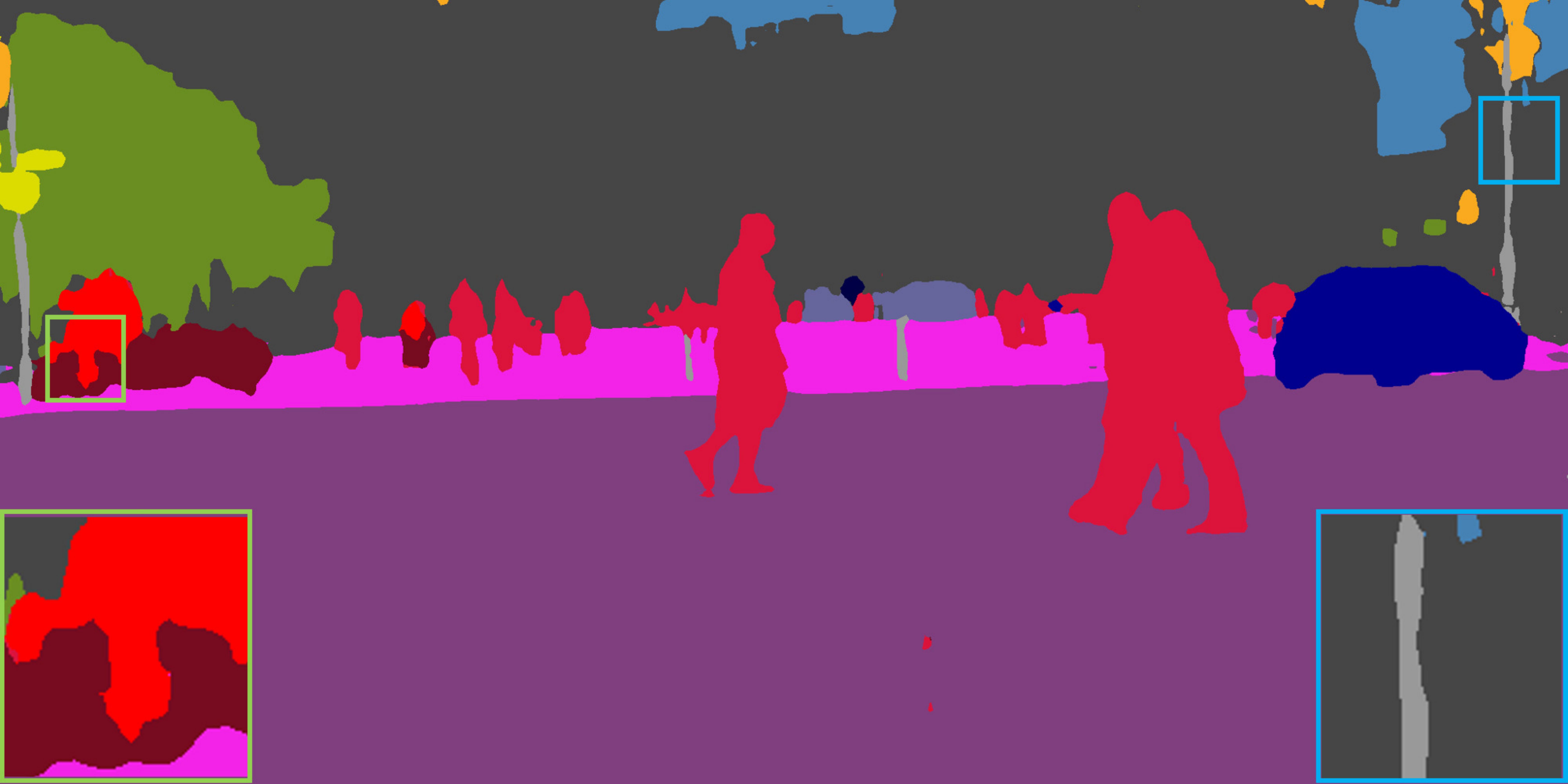}\\
		{\scriptsize (d) sub4 branch} & {\scriptsize (e) sub24 branch} & {\scriptsize (f) sub124 branch}\\
	\end{tabular}
	\caption{Visual prediction improvement of ICNet in each branch on Cityscapes dataset.}
	\label{fig:visualcityscapes}
\end{figure}

\begin{figure}[bpt]
	\centering
	\begin{tabular}{ccc}
		\includegraphics[width=0.32\linewidth]{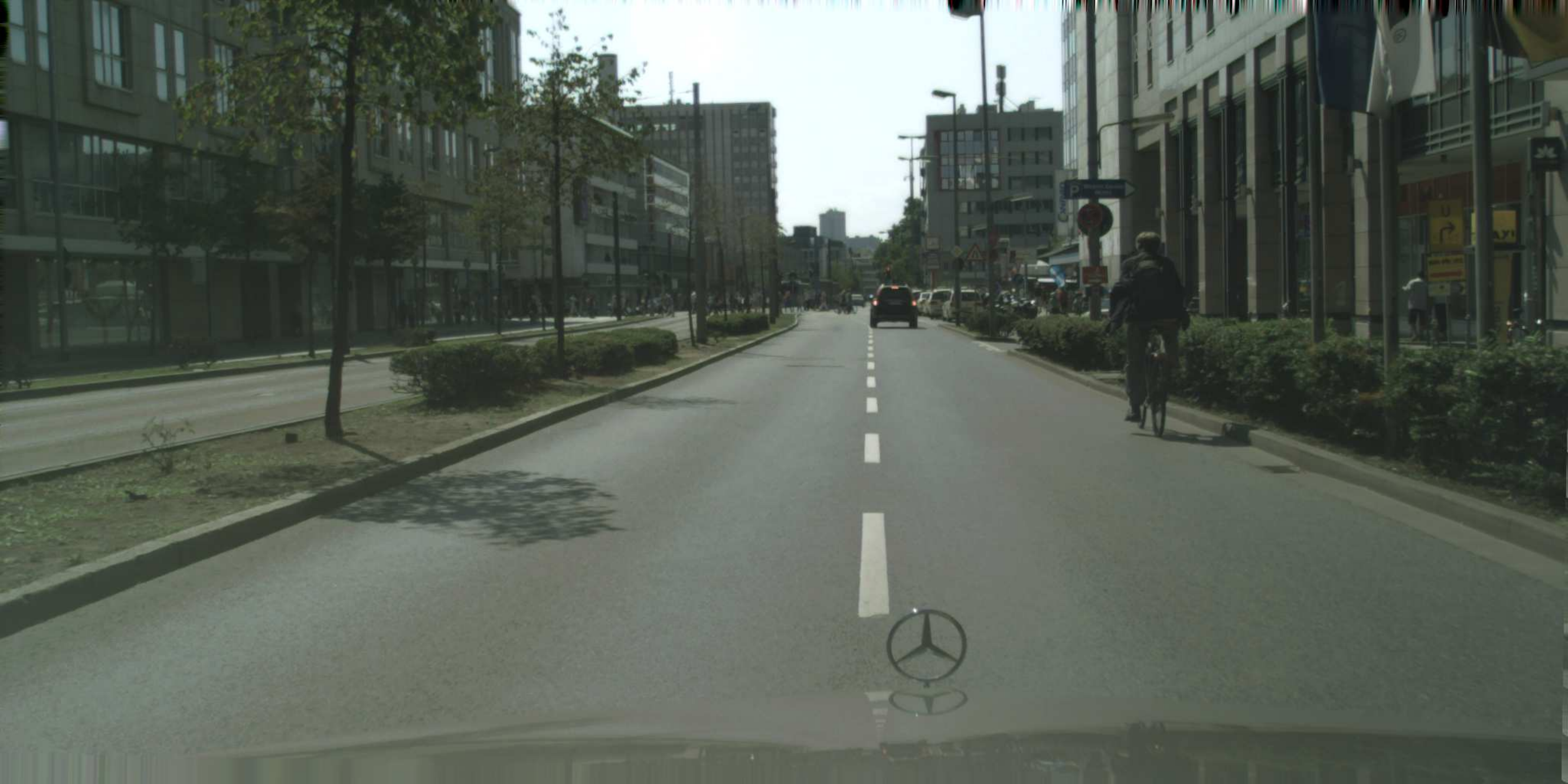}&
		\includegraphics[width=0.32\linewidth]{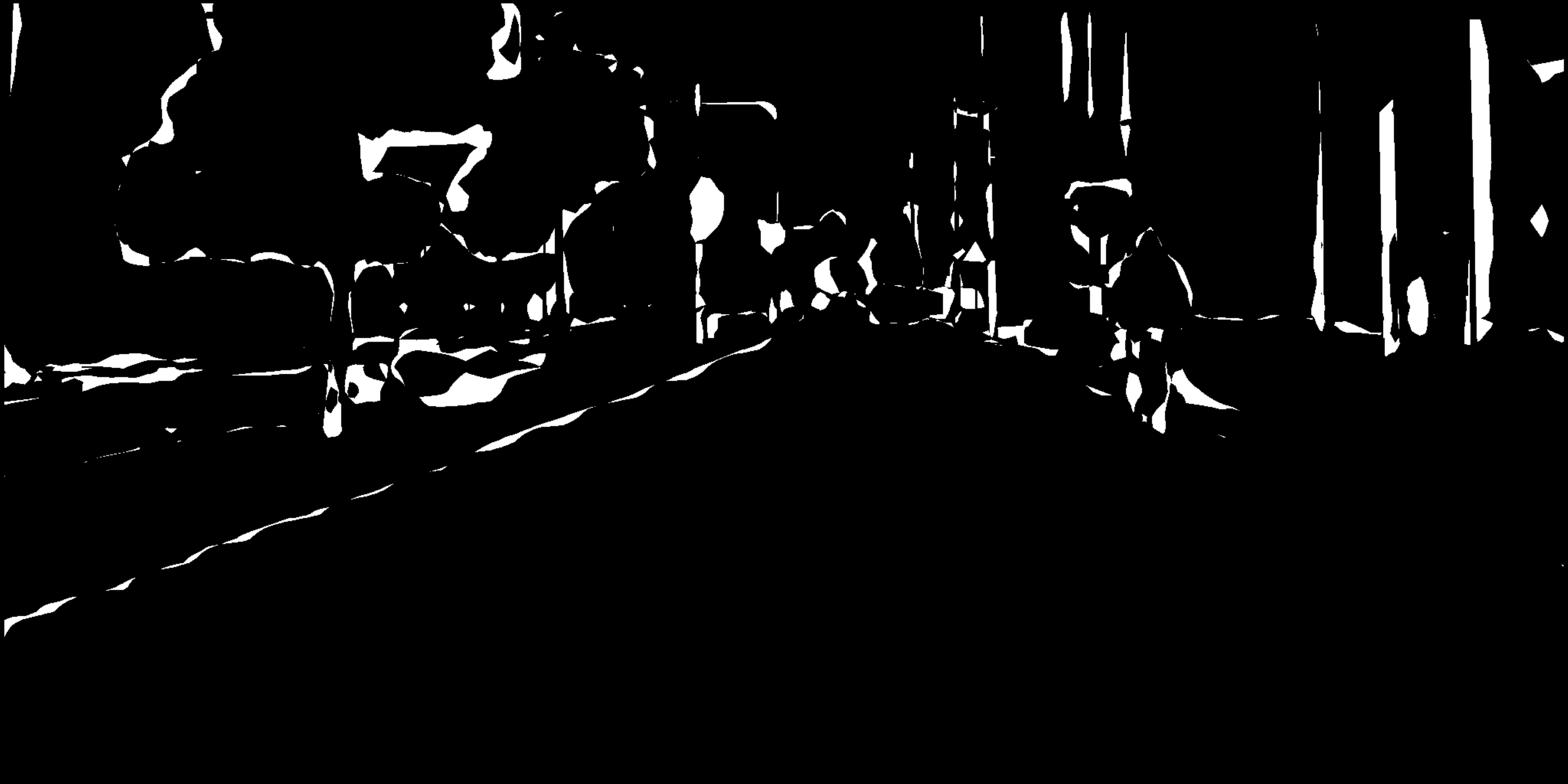}&
		\includegraphics[width=0.32\linewidth]{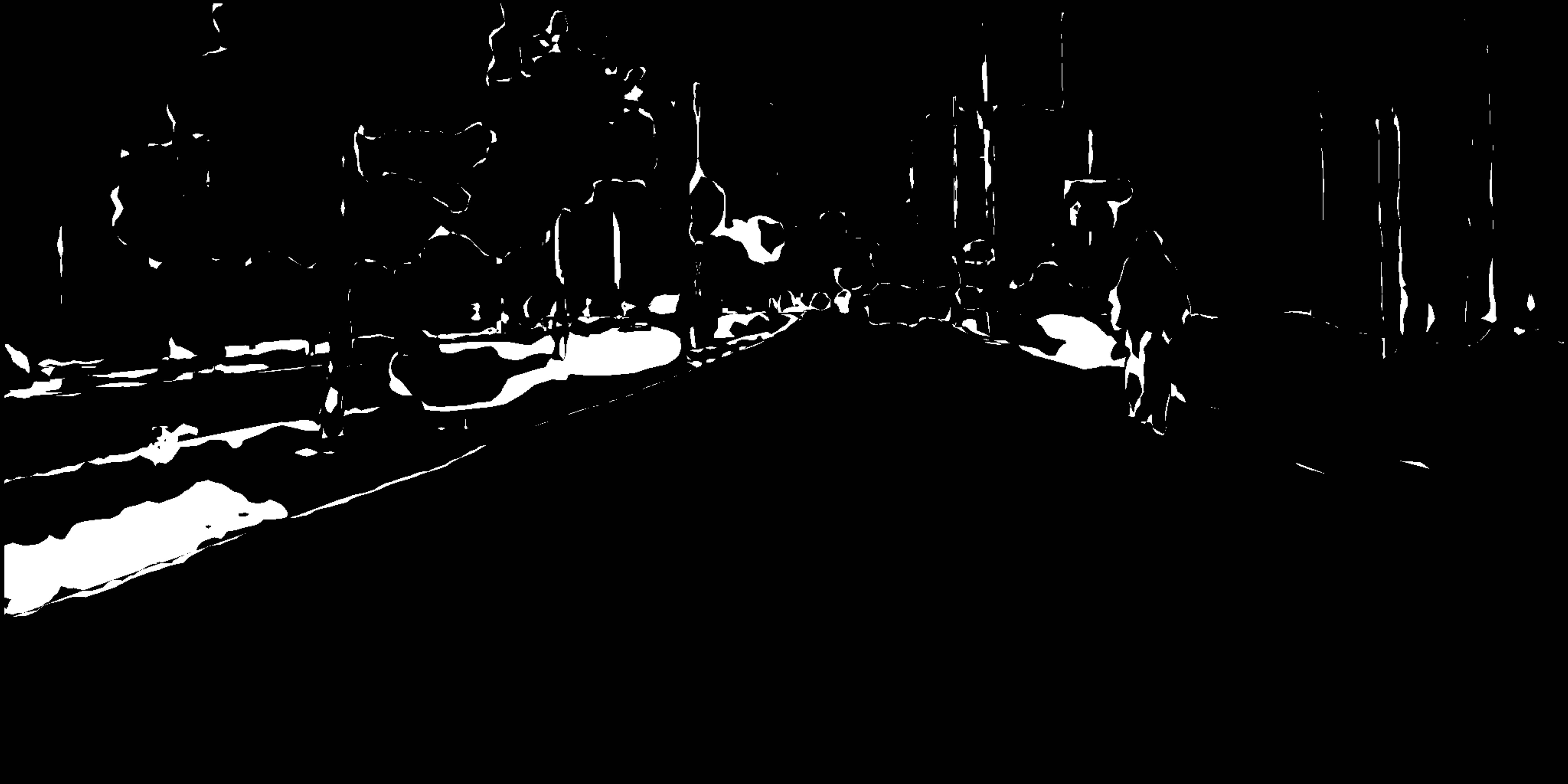}\\
		{\scriptsize (a) input image} & {\scriptsize (b) diff1} & {\scriptsize (c) diff2}\\
		\includegraphics[width=0.32\linewidth]{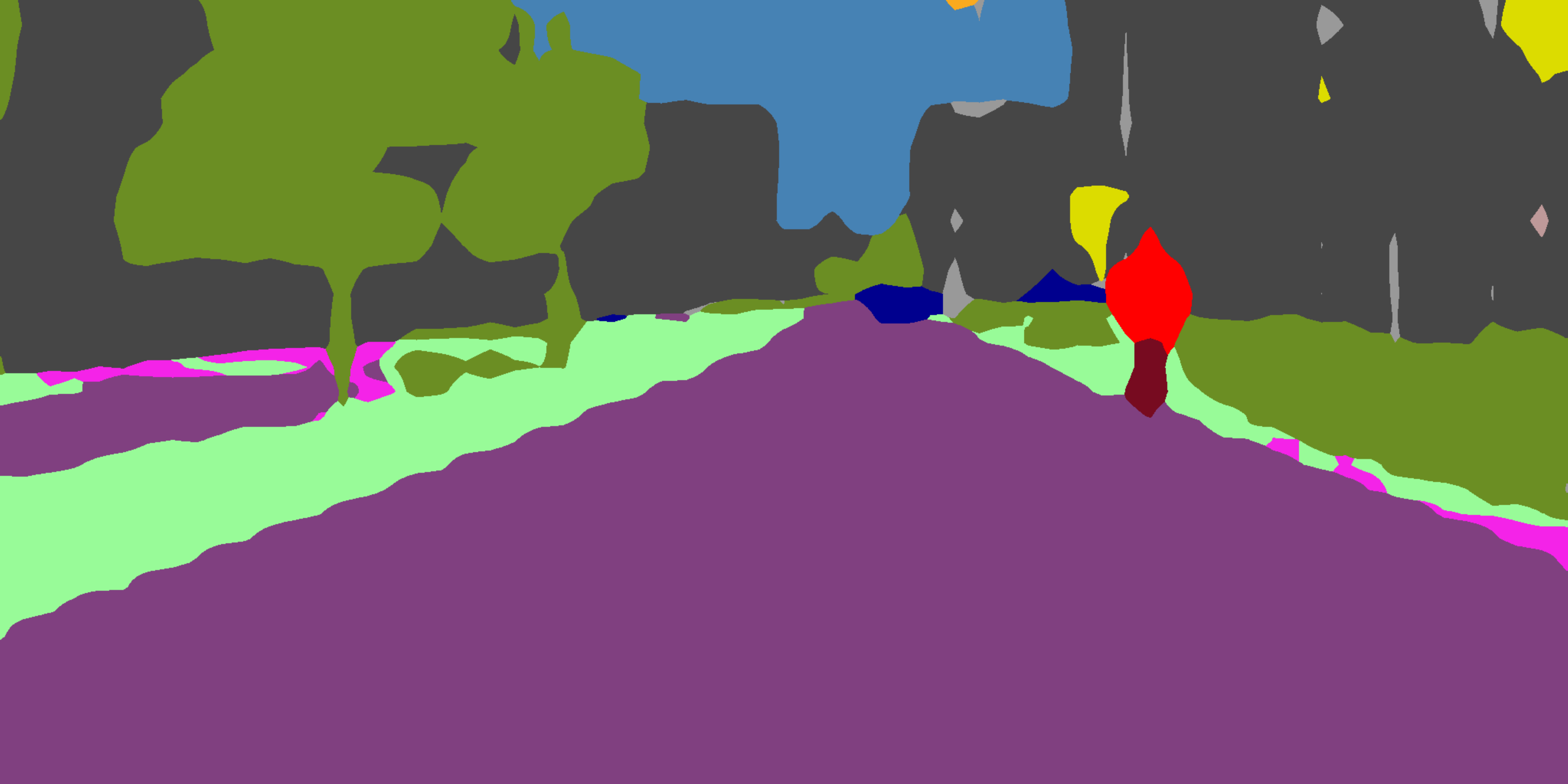}&
		\includegraphics[width=0.32\linewidth]{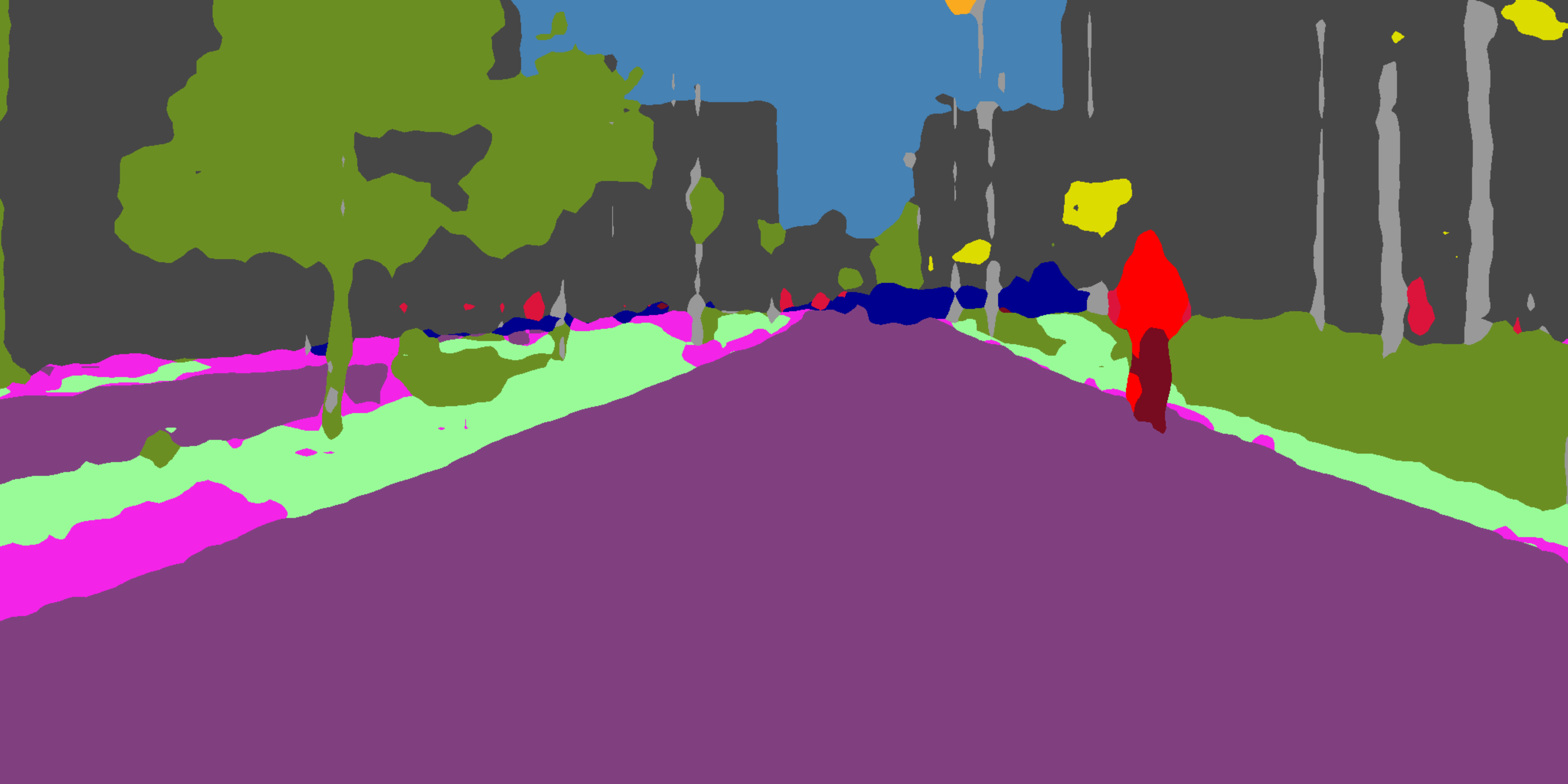}&
		\includegraphics[width=0.32\linewidth]{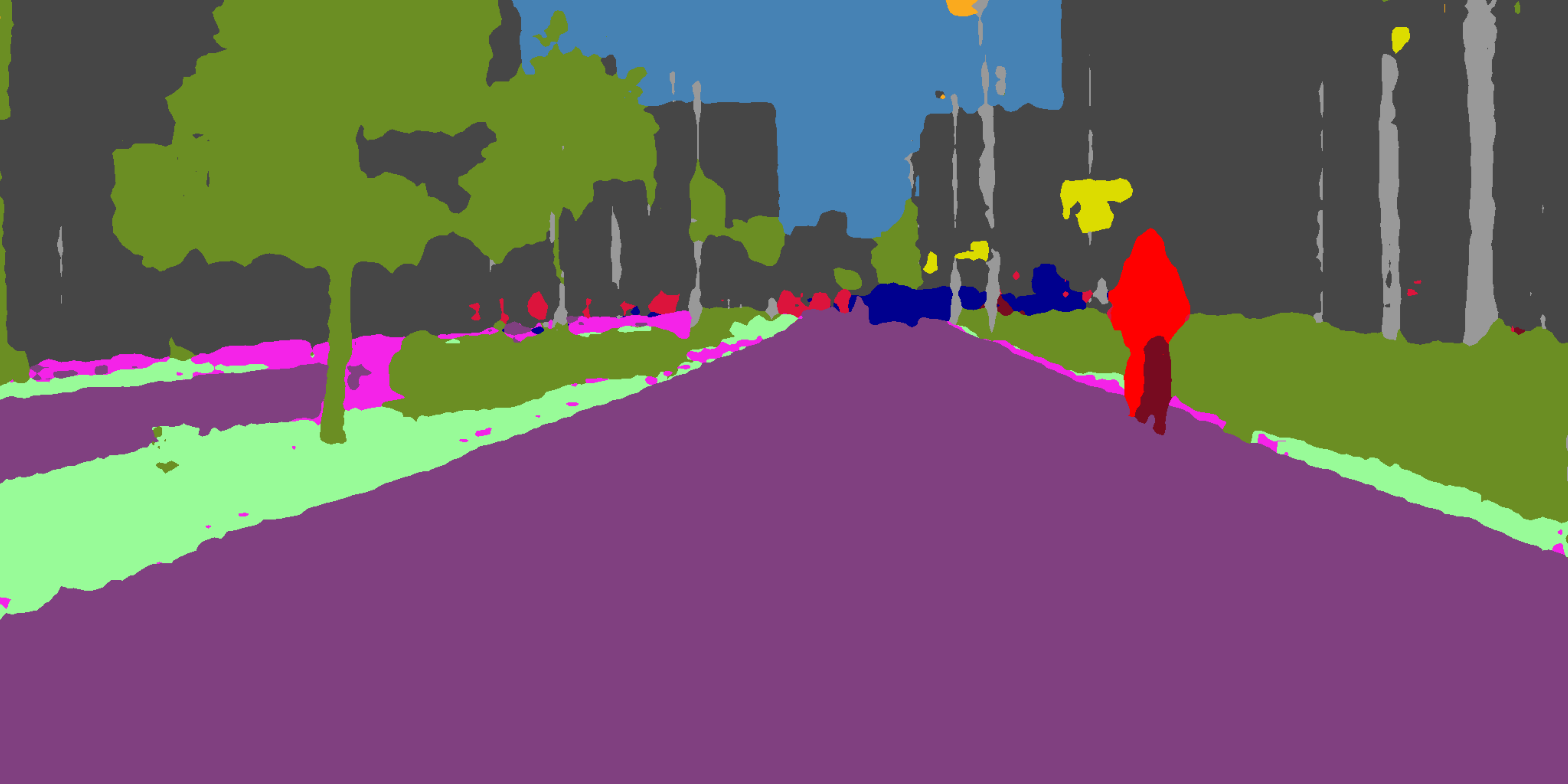}\\
		{\scriptsize (d) sub4 branch} & {\scriptsize (e) sub24 branch} & {\scriptsize (f) sub124 branch}\\
	\end{tabular}
	\caption{Visual prediction improvement of ICNet. White regions in `diff1' and `diff2' denote prediction difference between `sub24' and `sub4', and between `sub124' and `sub24' respectively.}
	\label{fig:cascadeprediction}
\end{figure}

\subsubsection{Visual Improvement}
Figs.~\ref{fig:visualcityscapes} and \ref{fig:cascadeprediction} show the visual results of ICNet on Cityscapes. With proposed gradual feature fusion steps and cascade label guidance structure, we produce decent prediction results. Intriguingly, output of the `sub4' branch can already capture most of semantically meaningful objects. But the prediction is coarse due to the low-resolution input. It misses a few small-size important regions, such as poles and traffic signs. 

With the help of medium-resolution information, many of these regions are re-estimated and recovered as shown in the `sub24' branch. It is noticeable that objects far from the camera, such as a few persons, are still missing with blurry object boundaries. The `sub124' branch with full-resolution input helps refine these details -- the output of this branch is undoubted the best. It manifests that our different-resolution information is properly made use of in this framework.
%We include more video examples in our supplementary video.

\begin{figure}[bpt]
	\begin{center}
		\includegraphics[width=0.85\linewidth]{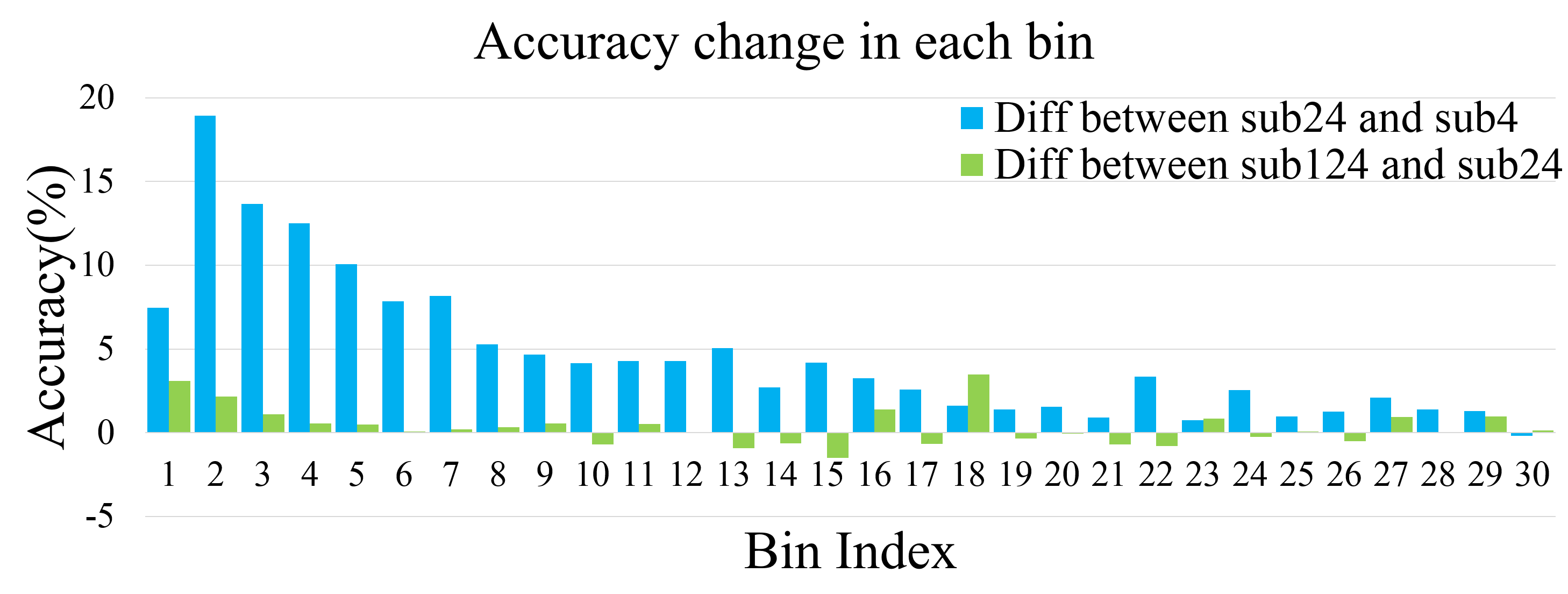}%{figure/binacc.eps}
	\end{center}
	\caption{Quantitative analysis of accuracy change in connected components.}
	\label{fig:binacc}
\end{figure}

\begin{table}[bpt]
	\begin{minipage}[t]{0.49\linewidth}
		\centering
		\caption{Results on CamVid test set with time reported on resolution $720 \times 960$.}
		\setlength{\tabcolsep}{5pt}
		\label{tab:camvid}
		\scriptsize
		\begin{tabular}{l | c c c}
			\toprule[1pt]
			Method & mIoU & Time & Frame \\
			& (\%) & (ms) & {fps} \\
			\hline
			\hline
			SegNet~\cite{badrinarayanan2015segnet} & 46.4 & 217 & 4.6 \\
			DPN~\cite{liu2015semantic} & 60.1 & 830 & 1.2 \\
			DeepLab~\cite{chen2015semantic} & 61.6 & 203 & 4.9 \\
			Dilation8~\cite{yu2016multi} & 65.3 & 227 & 4.4 \\
			PSPNet50~\cite{zhao2017pspnet} & 69.1 & 185 & 5.4 \\
			\hline
			\hline
			\rowcolor{Gray}
			ICNet & 67.1 & 36 & 27.8 \\
			\bottomrule[1pt]
		\end{tabular}
		
	\end{minipage}
	\hfill
	\begin{minipage}[t]{0.49\linewidth}
		\centering
		\caption{Results on COCO-Stuff test set with time reported on resolution $640 \times 640$.}
		\setlength{\tabcolsep}{5pt}
		\label{tab:cocostuff}
		\scriptsize
		\begin{tabular}{l | c c c}
			\toprule[1pt]
			Method & mIoU & Time & Frame \\
			& (\%) & (ms) & {fps} \\
			\hline
			\hline
			FCN~\cite{long2015fully} & 22.7 & 169 & 5.9 \\
			DeepLab~\cite{chen2015semantic} & 26.9 & 124 & 8.1 \\
			PSPNet50~\cite{zhao2017pspnet} & 32.6 & 151 & 6.6 \\
			\hline
			\hline
			\rowcolor{Gray}
			ICNet & 29.1 & 28 & 35.7 \\
			\bottomrule[1pt]
		\end{tabular}
	\end{minipage}
\end{table}

\iffalse
\begin{figure}[bpt]
	\centering
	\includegraphics[width=0.99\linewidth]{figure/camvidvisual.eps}
	\caption{Visual results of ICNet on CamVid dataset.}
	\label{fig:camvidvisual}
\end{figure}
\fi

\subsubsection{Quantitative Analysis}
To further understand accuracy gain in each branch, we quantitatively analyze the predicted label maps based on connected components. For each connected region $R_{i}$, we calculate the number of pixels it contains, denoted as $S_{i}$. Then we count the number of pixels correctly predicted in the corresponding map as $s_{i}$. The predicted region accuracy $p_{i}$ in $R_{i}$ is thus ${s_{i}}/{S_{i}}$. According to the region size $S_{i}$, we project these regions onto a histogram $\mathcal{H}$ with interval $\mathcal{K}$ and average all related region accuracy $p_{i}$ as the value of current bin.

In experiments, we set bin size of the histogram as 30 and interval $\mathcal{K}$ as 3,000. It thus covers region size $S_{i}$ between 1 to 90K. We ignore regions with size exceeding 90K. Fig.~\ref{fig:binacc} shows the accuracy change in each bin. The blue histogram stands for the difference between `sub24' and `sub4' while the green histogram shows the difference between `sub124' and `sub24'. For both histograms, the large difference is mainly on the front bins with small region sizes. This manifests that small region objects like traffic light and pole can be well improved in our framework. The front changes are large positives, proving that `sub24' can restore much information on small objects on top of `sub4'. `sub124' is also very useful compared to `sub24'.

\subsection{CamVid}
CamVid~\cite{BrostowFC09} dataset contains images extracted from high resolution video sequences with resolution up to $720 \times 960$. For easy comparison with prior work, we adopt the split of Sturgess et al.~\cite{sturgess2009combining}, which partitions the dataset into 367, 100, and 233 images for training, validation and testing respectively. 11 semantic classes are used for evaluation.

The testing results are listed in Table~\ref{tab:camvid}, our base-model is no compressed PSPNet50. ICNet gets much faster inference speed than other methods on this high resolution, reaching the real-time speed of 27.8 fps, 5.7 times faster than the second one and 5.1 times faster compared to the basic model. Apart from high efficiency, it also accomplishes high quality segmentation. Visual results are provided in the supplementary material.
%Visual results are shown in Fig.~\ref{fig:camvidvisual}.

\subsection{COCO-Stuff}
COCO-Stuff~\cite{caesar2016coco} is a recently labeled dataset based on MS-COCO~\cite{lin2014microsoft} for stuff segmentation in context. We evaluate ICNet following the split in \cite{caesar2016coco} that 9K images are used for training and another 1K for testing. This dataset is much more complex for multiple categories -- up to 182 classes are used for evaluation, including 91 thing and 91 stuff classes.

Table~\ref{tab:cocostuff} shows the testing results. ICNet still performs satisfyingly regarding common thing and stuff understanding. It is more efficient and accurate than modern segmentation frameworks, such as FCN and DeepLab. Compared to our baseline model, it achieves 5.4 times speedup. Visual predictions are provided in the supplementary material.
%Visual prediction results are shown in Fig.~\ref{fig:cocostuffvisual}.

\iffalse
\begin{figure}[bpt]
	\centering
	\includegraphics[width=0.99\linewidth]{figure/cocostuffvisual.eps}
	\caption{Visual results of ICNet on COCO-Stuff dataset.}
	\label{fig:cocostuffvisual}
\end{figure}
\fi

%------------------------------------------------------------------------
\section{Conclusion}
We have proposed a real-time semantic segmentation system ICNet. It incorporates effective strategies to accelerate network inference speed without sacrificing much performance. The major contributions include the new framework for saving operations in multiple resolutions and the powerful fusion unit.

We believe the optimal balance of speed and accuracy makes our system important since it can benefit many other tasks that require fast scene and object segmentation. It greatly enhances the practicality of semantic segmentation in other disciplines.

\clearpage

\bibliographystyle{splncs}
\bibliography{egbib}

\begin{thebibliography}{1}

\bibitem{chen2017rethinking}
Chen, L.C., Papandreou, G., Schroff, F., Adam, H.:
\newblock Rethinking atrous convolution for semantic image segmentation.
\newblock arXiv:1706.05587 (2017)

\bibitem{pohlen2017FRRN}
Pohlen, T., Hermans, A., Mathias, M., Leibe, B.:
\newblock Full-resolution residual networks for semantic segmentation in street
  scenes.
\newblock In: CVPR. (2017)

\bibitem{lin2017refine}
Lin, G., Milan, A., Shen, C., Reid, I.D.:
\newblock Refinenet: Multi-path refinement networks for high-resolution
  semantic segmentation.
\newblock In: CVPR. (2017)

\bibitem{wang2017duc}
Wang, P., Chen, P., Yuan, Y., Liu, D., Huang, Z., Hou, X., Cottrell, G.W.:
\newblock Understanding convolution for semantic segmentation.
\newblock In: WACV. (2018)

\bibitem{wu2016wider}
Wu, Z., Shen, C., van~den Hengel, A.:
\newblock Wider or deeper: Revisiting the resnet model for visual recognition.
\newblock arXiv:1611.10080 (2016)

\bibitem{zhao2017pspnet}
Zhao, H., Shi, J., Qi, X., Wang, X., Jia, J.:
\newblock Pyramid scene parsing network.
\newblock In: CVPR. (2017)

\end{thebibliography}


\begin{thebibliography}{10}

\bibitem{long2015fully}
Long, J., Shelhamer, E., Darrell, T.:
\newblock Fully convolutional networks for semantic segmentation.
\newblock In: CVPR. (2015)

\bibitem{chen2015semantic}
Chen, L., Papandreou, G., Kokkinos, I., Murphy, K., Yuille, A.L.:
\newblock Semantic image segmentation with deep convolutional nets and fully
  connected crfs.
\newblock ICLR (2015)

\bibitem{badrinarayanan2015segnet}
Badrinarayanan, V., Kendall, A., Cipolla, R.:
\newblock Segnet: A deep convolutional encoder-decoder architecture for image
  segmentation.
\newblock arXiv:1511.00561 (2015)

\bibitem{noh2015learning}
Noh, H., Hong, S., Han, B.:
\newblock Learning deconvolution network for semantic segmentation.
\newblock In: ICCV. (2015)

\bibitem{zhao2017pspnet}
Zhao, H., Shi, J., Qi, X., Wang, X., Jia, J.:
\newblock Pyramid scene parsing network.
\newblock In: CVPR. (2017)

\bibitem{wu2016wider}
Wu, Z., Shen, C., van~den Hengel, A.:
\newblock Wider or deeper: Revisiting the resnet model for visual recognition.
\newblock arXiv:1611.10080 (2016)

\bibitem{cordts2016cityscapes}
Cordts, M., Omran, M., Ramos, S., Rehfeld, T., Enzweiler, M., Benenson, R.,
  Franke, U., Roth, S., Schiele, B.:
\newblock The cityscapes dataset for semantic urban scene understanding.
\newblock In: CVPR. (2016)

\bibitem{paszke2016enet}
Paszke, A., Chaurasia, A., Kim, S., Culurciello, E.:
\newblock Enet: A deep neural network architecture for real-time semantic
  segmentation.
\newblock arXiv:1606.02147 (2016)

\bibitem{treml2016speeding}
Treml, M., Arjona-Medina, J., Unterthiner, T., Durgesh, R., Friedmann, F.,
  Schuberth, P., Mayr, A., Heusel, M., Hofmarcher, M., Widrich, M., Nessler1,
  B., Hochreiter, S.:
\newblock Speeding up semantic segmentation for autonomous driving.
\newblock NIPS Workshop (2016)

\bibitem{wang2017duc}
Wang, P., Chen, P., Yuan, Y., Liu, D., Huang, Z., Hou, X., Cottrell, G.W.:
\newblock Understanding convolution for semantic segmentation.
\newblock arXiv:1702.08502 (2017)

\bibitem{lin2017refine}
Lin, G., Milan, A., Shen, C., Reid, I.D.:
\newblock Refinenet: Multi-path refinement networks for high-resolution
  semantic segmentation.
\newblock In: CVPR. (2017)

\bibitem{pohlen2017FRRN}
Pohlen, T., Hermans, A., Mathias, M., Leibe, B.:
\newblock Full-resolution residual networks for semantic segmentation in street
  scenes.
\newblock In: CVPR. (2017)

\bibitem{chen2016deeplab}
Chen, L., Papandreou, G., Kokkinos, I., Murphy, K., Yuille, A.L.:
\newblock Deeplab: Semantic image segmentation with deep convolutional nets,
  atrous convolution, and fully connected crfs.
\newblock arXiv:1606.00915 (2016)

\bibitem{yu2016multi}
Yu, F., Koltun, V.:
\newblock Multi-scale context aggregation by dilated convolutions.
\newblock ICLR (2016)

\bibitem{liu2015semantic}
Liu, Z., Li, X., Luo, P., Loy, C.C., Tang, X.:
\newblock Semantic image segmentation via deep parsing network.
\newblock In: ICCV. (2015)

\bibitem{zheng2015conditional}
Zheng, S., Jayasumana, S., Romera{-}Paredes, B., Vineet, V., Su, Z., Du, D.,
  Huang, C., Torr, P.H.S.:
\newblock Conditional random fields as recurrent neural networks.
\newblock In: ICCV. (2015)

\bibitem{BrostowFC09}
Brostow, G.J., Fauqueur, J., Cipolla, R.:
\newblock Semantic object classes in video: {A} high-definition ground truth
  database.
\newblock Pattern Recognition Letters (2009)

\bibitem{caesar2016coco}
Caesar, H., Uijlings, J., Ferrari, V.:
\newblock Coco-stuff: Thing and stuff classes in context.
\newblock arXiv:1612.03716 (2016)

\bibitem{liu2011nonparametric}
Liu, C., Yuen, J., Torralba, A.:
\newblock Nonparametric scene parsing via label transfer.
\newblock TPAMI (2011)

\bibitem{chen2015attention}
Chen, L., Yang, Y., Wang, J., Xu, W., Yuille, A.L.:
\newblock Attention to scale: Scale-aware semantic image segmentation.
\newblock In: CVPR. (2016)

\bibitem{hariharan2015hypercolumns}
Hariharan, B., Arbel{\'{a}}ez, P.A., Girshick, R.B., Malik, J.:
\newblock Hypercolumns for object segmentation and fine-grained localization.
\newblock In: CVPR. (2015)

\bibitem{xia2016zoom}
Xia, F., Wang, P., Chen, L., Yuille, A.L.:
\newblock Zoom better to see clearer: Human and object parsing with
  hierarchical auto-zoom net.
\newblock In: ECCV. (2016)

\bibitem{girshick2015fastrcnn}
Girshick, R.:
\newblock Fast {R-CNN}.
\newblock In: ICCV. (2015)

\bibitem{ren2015faster}
Ren, S., He, K., Girshick, R., Sun, J.:
\newblock Faster {R-CNN}: Towards real-time object detection with region
  proposal networks.
\newblock In: NIPS. (2015)

\bibitem{Redmon2016yolo}
Redmon, J., Divvala, S.K., Girshick, R.B., Farhadi, A.:
\newblock You only look once: Unified, real-time object detection.
\newblock In: CVPR. (2016)

\bibitem{Redmon2017yolo2}
Redmon, J., Farhadi, A.:
\newblock {YOLO9000:} better, faster, stronger.
\newblock In: CVPR. (2017)

\bibitem{liu2016ssd}
Liu, W., Anguelov, D., Erhan, D., Szegedy, C., Reed, S.E., Fu, C., Berg, A.C.:
\newblock Ssd: Single shot multibox detector.
\newblock In: ECCV. (2016)

\bibitem{romera2017efficient}
Romera, E., Alvarez, J.M., Bergasa, L.M., Arroyo, R.:
\newblock Efficient convnet for real-time semantic segmentation.
\newblock In: Intelligent Vehicles Symposium (IV). (2017)

\bibitem{Shelhamer2016clockfcn}
Shelhamer, E., Rakelly, K., Hoffman, J., Darrell, T.:
\newblock Clockwork convnets for video semantic segmentation.
\newblock In: ECCV Workshop. (2016)

\bibitem{zhu2017dff}
Zhu, X., Xiong, Y., Dai, J., Yuan, L., Wei, Y.:
\newblock Deep feature flow for video recognition.
\newblock In: CVPR. (2017)

\bibitem{kundu2016feature}
Kundu, A., Vineet, V., Koltun, V.:
\newblock Feature space optimization for semantic video segmentation.
\newblock In: CVPR. (2016)

\bibitem{gadde2017semantic}
Gadde, R., Jampani, V., Gehler, P.V.:
\newblock Semantic video cnns through representation warping.
\newblock In: ICCV. (2017)

\bibitem{ronneberger2015unet}
Ronneberger, O., Fischer, P., Brox, T.:
\newblock U-net: Convolutional networks for biomedical image segmentation.
\newblock In: MICCAI. (2015)

\bibitem{ghiasi2016laplacian}
Ghiasi, G., Fowlkes, C.C.:
\newblock Laplacian pyramid reconstruction and refinement for semantic
  segmentation.
\newblock In: ECCV. (2016)

\bibitem{everingham2010pascal}
Everingham, M., Gool, L.J.V., Williams, C.K.I., Winn, J.M., Zisserman, A.:
\newblock The pascal visual object classes {VOC} challenge.
\newblock IJCV (2010)

\bibitem{zhou2016semantic}
Zhou, B., Zhao, H., Puig, X., Fidler, S., Barriuso, A., Torralba, A.:
\newblock Semantic understanding of scenes through the {ADE20K} dataset.
\newblock arXiv:1608.05442 (2016)

\bibitem{jia2014caffe}
Jia, Y., Shelhamer, E., Donahue, J., Karayev, S., Long, J., Girshick, R.B.,
  Guadarrama, S., Darrell, T.:
\newblock Caffe: Convolutional architecture for fast feature embedding.
\newblock In: ACM MM. (2014)

\bibitem{Iandola2016squeezenet}
Iandola, F.N., Moskewicz, M.W., Ashraf, K., Han, S., Dally, W.J., Keutzer, K.:
\newblock Squeezenet: Alexnet-level accuracy with 50x fewer parameters and
  {\textless}1mb model size.
\newblock arXiv:1602.07360 (2016)

\bibitem{han2016deepcom}
Han, S., Mao, H., Dally, W.J.:
\newblock Deep compression: Compressing deep neural network with pruning,
  trained quantization and huffman coding.
\newblock In: ICLR. (2016)

\bibitem{han2017dsd}
Han, S., Pool, J., Narang, S., Mao, H., Tang, S., Elsen, E., Catanzaro, B.,
  Tran, J., Dally, W.J.:
\newblock {DSD:} regularizing deep neural networks with dense-sparse-dense
  training flow.
\newblock In: ICLR. (2017)

\bibitem{li2017pruning}
Li, H., Kadav, A., Durdanovic, I., Samet, H., Graf, H.P.:
\newblock Pruning filters for efficient convnets.
\newblock In: ICLR. (2017)

\bibitem{sturgess2009combining}
Sturgess, P., Alahari, K., Ladicky, L., Torr, P.H.:
\newblock Combining appearance and structure from motion features for road
  scene understanding.
\newblock In: BMVC. (2009)

\bibitem{lin2014microsoft}
Lin, T., Maire, M., Belongie, S.J., Hays, J., Perona, P., Ramanan, D.,
  Doll{\'{a}}r, P., Zitnick, C.L.:
\newblock Microsoft coco: Common objects in context.
\newblock In: ECCV. (2014)

\end{thebibliography}
%\bibliography{../../cvbib/egbib}
\end{document}